\documentclass[10pt,twocolumn,letterpaper]{article}

\usepackage{iccv}
\usepackage{times}
\usepackage{epsfig}
\usepackage{caption}
\usepackage{graphicx}
\usepackage{amsmath}
\usepackage{amssymb}
\usepackage{balance}
\usepackage{color}
\usepackage{float}
\usepackage{pifont}%
\usepackage[printonlyused]{acronym}
\usepackage{authblk}
\usepackage{booktabs}
\usepackage[numbers,sort]{natbib}

\newcommand{\name}{ToFu\xspace}
\newcommand{\longname}{\textbf{To}pologically consistent \textbf{F}ace from m\textbf{u}lti-view\xspace}
\newcommand{\cmark}{\ding{51}}%
\newcommand{\xmark}{\ding{55}}%
\newcommand{\qheading}[1]{\noindent\textbf{#1}}
\usepackage[ruled,vlined]{algorithm2e}
\newcommand\norm[1]{\left\lVert#1\right\rVert}

\acrodef{ICP}{Iterative Closest Point}
\acrodef{MVS}{Multi-View Stereo}

\usepackage[pagebackref=true,breaklinks=true,colorlinks,bookmarks=false]{hyperref}

\iccvfinalcopy %

\begin{document}

\title{
Topologically Consistent Multi-View Face Inference Using Volumetric Sampling
}

\makeatletter
\renewcommand\AB@affilsepx{, \protect\Affilfont}
\makeatother

\author[1,2]{Tianye Li}
\author[1,2]{Shichen Liu}
\author[3]{Timo Bolkart}
\author[1,2]{Jiayi Liu}
\author[1,2]{Hao Li}
\author[1]{Yajie Zhao}
\affil[1]{USC Institute for Creative Technologies}
\affil[2]{USC}
\affil[3]{MPI for Intelligent Systems, T\"ubingen}

\twocolumn[{%
	\renewcommand\twocolumn[1][]{#1}%
	\maketitle
	\begin{center}
		\centering
        \includegraphics[width=0.99\linewidth]{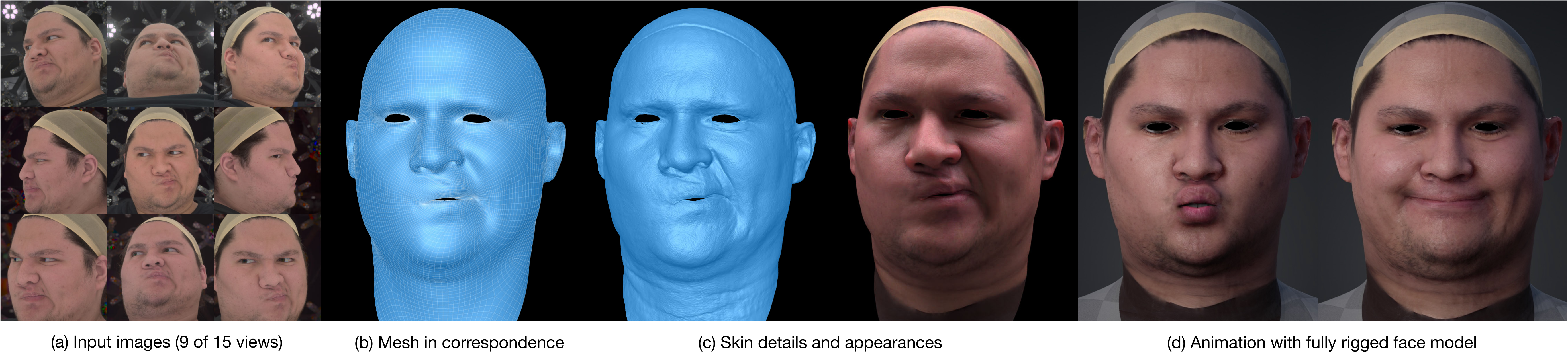}
		\captionof{figure}{
		  Given (a) multi-view images, our face modeling framework \name uses volumetric sampling to predict (b) accurate base meshes in consistent topology as well as (c) high-resolution details and appearances.
		  Our efficient pipeline enables (d) rapid creation of production-quality avatars for animation.
		}
		\label{fig:teaser}
	\end{center}%
}]

\begin{abstract}
High-fidelity face digitization solutions often combine multi-view stereo (MVS) techniques for 3D reconstruction and a non-rigid registration step to establish dense correspondence across identities and expressions.  
A common problem is the need for manual clean-up after the MVS step, as 3D scans are typically affected by noise and outliers and contain hairy surface regions that need to be cleaned up by artists.
Furthermore, mesh registration tends to fail for extreme facial expressions.
Most learning-based methods use an underlying 3D morphable model (3DMM) to ensure robustness, but this limits the output accuracy for extreme facial expressions.
In addition, the global bottleneck of regression architectures cannot produce meshes that tightly fit the ground truth surfaces.
We propose \name, \longname, a geometry inference framework that can produce topologically consistent meshes across facial identities and expressions using a volumetric representation instead of an explicit underlying 3DMM.
Our novel progressive mesh generation network embeds the topological structure of the face in a feature volume, sampled from geometry-aware local features.
A coarse-to-fine architecture facilitates dense and accurate facial mesh predictions in a consistent mesh topology.
\name further captures displacement maps for pore-level geometric details and facilitates high-quality rendering in the form of albedo and specular reflectance maps. 
These high-quality assets are readily usable by production studios for avatar creation, animation and physically-based skin rendering.
We demonstrate state-of-the-art geometric and correspondence accuracy, while only taking 0.385 seconds to compute a mesh with 10K vertices, which is three orders of magnitude faster than traditional techniques.
The code and the model are available for research purposes at \url{https://tianyeli.github.io/tofu}.
\end{abstract}

\section{Introduction}
\label{sec:intro}

Creating high-fidelity digital humans is not only highly sought after in the film and gaming industry, but is also gaining interest in consumer applications, ranging from telepresence in AR/VR to virtual fashion models and virtual assistants.
While fully automated single-view avatar digitization solutions exist ~\cite{Hu_2017_SIGGRAPH, nagano2018pagan, ichim2015dynamic,tewari17MoFA, wu2019mvf}, professional studios still opt for high resolution multi-view images as input, to ensure the highest possible fidelity and surface coverage in a controlled setting~\cite{Ghosh_2011_SIGGRAPH, Beeler2011, seol2016creating, lombardi2019neuralvolume, riviere2020single, ma2007rapid, gotardo2018practical} instead of unconstrained input data. 
Typically, high-resolution geometric details (${< 1mm} $ error) are desired along with high resolution physically-based material properties (at least $4K$).
Furthermore, to build a fully rigged face model for animation, a large number of facial scans and alignments (often over 30) are performed, typically following some conventions based on the Facial Action Coding System (FACS).

A typical approach used in production consists of using a multi-view stereo acquisition process to capture detailed 3D scans of each facial expression, and a non-rigid registration~\cite{li2009robust, Beeler2011} or inference method~\cite{li2020dynamic} is used to warp a 3D face model to each scan in order to ensure consistent mesh topology. Between these two steps, manual clean-up is often necessary to remove artifacts and unwanted surface regions, especially those with facial hair (beards, eyebrows) as well as teeth and neck regions. The registration process is often assisted with manual labeling tasks for correspondences and parameter tweaking to ensure accurate fitting. In a production setting, a completed rig of a person can easily take up to a week to finalize.

Several recent techniques have been introduced to automate this process by fitting a 3D model directly to a calibrated set of input images. The multi-view stereo face modeling method of~\cite{Fyff:2017} is not only particularly slow, but relies on dynamic sequences and carefully tuned parameters for each subject to ensure consistent parameterization between expressions. In particular facial expressions that are not captured continuously cannot ensure accurate topological consistencies. More recent deep learning approaches~\cite{wu2019mvf, bai2020dfnrmvs} use a 3D morphable model (3DMM) inference to obtain a coarse initial facial expression, but require a refinement step based on optimization to improve fitting accuracy.
Those methods are limited in fitting extreme expressions due to the constraints of linear 3DMMs and fitting tightly to the ground-truth face surfaces due to the global nature of their regression architectures.
The additional photometric refinement also tends to fit unwanted regions like facial hair.

We introduce a new volumetric approach for consistent 3D face mesh inference using multi-view images. Instead of relying explicitly on a mesh-based face model such as 3DMM, our volumetric approach is more general, allowing it to capture a wider range of expressions and subtle deformation details on the face. Our method is also three orders of magnitude faster than conventional methods, taking only 0.385 seconds to generate a dense 3D mesh (10K vertices) as well as produce additional assets for high-fidelity production use cases, such as albedo, specular, and high-resolution displacement maps.

To this end, we propose a progressive mesh generation network that can infer a topologically consistent mesh directly. Our volumetric architecture predicts vertex locations as probability distributions, along with volumetric features that are extracted using the underlying multi-view geometry. The topological structure of the face is embedded into this architecture using a hierarchical mesh representation and coarse-to-fine network. 

Our experiments show that \name is capable of producing highly accurate geometry consistent with topology automatically, while existing methods either rely on manual clean-up and parameter tuning, or are less accurate especially for subjects with facial hair. 
Since we can ensure a consistent parameterization across facial identities and expressions without any human input, our solution is suitable for scaled digitization of high-fidelity facial avatars, 
We not only reduce the turn around time for production, but is also provide a critical solution for generating large facial datasets, which is often associated with excessive manual labor.
Our main contributions are:
\begin{itemize}
    \item A novel volumetric feature sampling and refinement model for topologically consistent 3D mesh reconstruction from multi-view images. 
    \item An appearance capture network to infer high-resolution skin details and appearance maps, which, combined with the base mesh, forms a complete package suitable for production in animation and photorealisitic rendering.
    \item We demonstrate state-of-the-art performance for combined geometry and correspondence accuracy, while achieving mesh inference at near interactive rates.
    \item Code and model are publicly available.
\end{itemize}

\begin{figure*}[ht!]
    \begin{center}
	\includegraphics[width=1\textwidth]{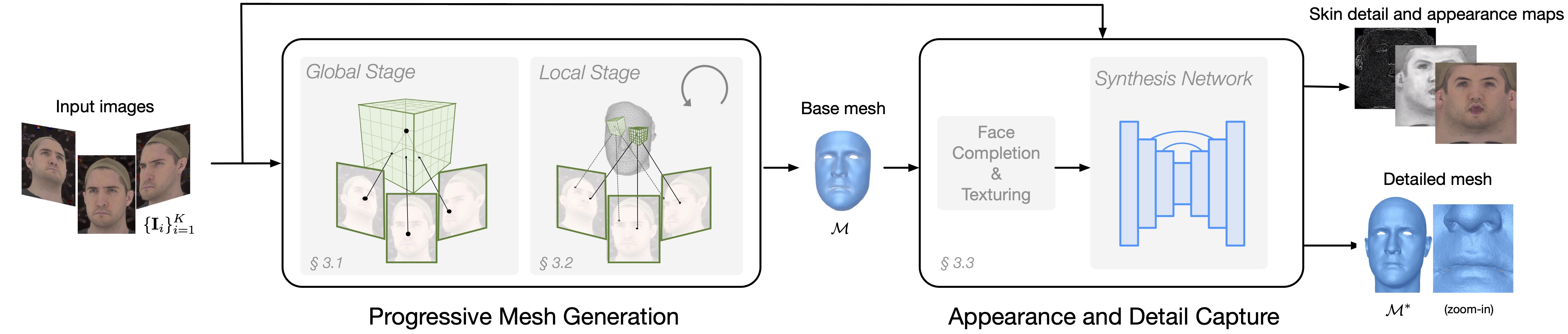}
    \end{center}
    \vspace{-10pt}
 	\caption{
    Overview of our end-to-end face modeling system. Given images captured from multi-views, the progressive mesh generation network predicts an accurate face mesh in consistent topology. Then the appearance and detail capture network synthesizes high-resolution skin detail and attribute maps, which enables highly detailed geometry and photo-realistic renderings.
	}
	  \label{fig:pipeline_system}
	  \vspace{-10pt}
\end{figure*}

\section{Related Work}
\label{sec:related_work}

\qheading{Face Capture.} 
Traditionally, face acquisition is separated into two steps, 3D face reconstruction and registration \cite{Egger2020}.
Facial geometry can be captured with laser scanners~\cite{levoy2000digital}, passive \ac{MVS} capture systems~\cite{Beeler_2010_SIGGRAPH}, dedicated active photometric stereo systems~\cite{ma2007rapid, Ghosh_2011_SIGGRAPH}, or depth sensors based on structured light or time-of-flight sensors.
Among these, \ac{MVS} is the most commonly used \cite{esteban2004silhouette,furukawa2008high,goesele2006multi,kolmogorov2002multi,pons2005modelling,vogiatzis2005multi}.
Although these approaches produce high-quality geometry, they  suffer from heavy computation due to the pairwise features matching across views, and they tend to fail in case of sparse view inputs due to the lack of overlapping neighboring views. 
More recently, deep neural networks learn multi-view feature matching for 3D geometry reconstruction \cite{gu2020casmvsnet,kar2017learning,im2019dpsnet,yao2018mvsnet, sitzmann2019deepvoxels}.
Compared to classical \ac{MVS} methods, these learning based methods represent a trade-off between accuracy and efficacy.
All these \ac{MVS} methods output unstructured meshes, while our method produces meshes in dense vertex correspondence. 

Most registration methods use a template mesh and fit it to the scan surface by minimizing the distance between the scan's surface and the template.
For optimization, the template mesh is commonly parameterized with a statistical shape space \cite{Amberg_2008,blanz2003reanimating,blanz1999morphable,RuilongLi2020} or a general blendshape basis \cite{Salazar2014}.
Other approaches directly optimize the vertices of the template mesh using a non-rigid \ac{ICP} \cite{li2009robust}, with a statistical model as regularizer \cite{FLAME2017}, or jointly optimize correspondence across an entire dataset in a groupwise fashion \cite{BolkartWuhrer2015ICCV,Zhang2016}.
For a more thorough review of face acquisition and registration, see Egger et al.~\cite{Egger2020}.
All these registration methods solve for facial correspondence independent from the data acquisition.
Therefore, errors in the raw scan data propagate into the registration.

Only few methods exist that are similar to our method of directly outputting high-quality registered 3D faces from calibrated multi-view input \cite{Beeler2011, Fyff:2017, borshukov2005universal, borshukov2003universal}.
While sharing a similar goal, our method goes beyond these approaches in several significant ways.
Unlike our method, they require calibrated multi-view image sequence input, contain multiple optimization steps (e.g. for building a subject specific template \cite{Fyff:2017}, or anchor frame meshes \cite{Beeler2011}), and are computationally slow (e.g. 25 minutes per frame for the coarse mesh reconstruction \cite{Fyff:2017}).
\name instead takes calibrated multi-view images as input (i.e. static) and directly outputs a high-quality mesh in dense vertex correspondence in $0.385$ seconds. 
Regardless, our method achieves stable reconstruction and registration results for sequence input.

\qheading{Model-based reconstruction.} 
A large body of work aims at reconstructing 3D faces from unconstrained images or monocular videos.
To constrain the problem, most methods estimate the coefficients of a statistical 3D morphable models (3DMM) in an optimization-based~\cite{aldrian2012inverse,bas2016fitting,blanz2002face,blanz1999morphable,Thies_2016_CVPR} or learning-based framework~\cite{Chang2018,Feng:SIGGRAPH:2021,Genova2018,richardson20163d, RingNet:CVPR:2019,tewari17MoFA,tran2018extreme}.
Due to the use of over-simplified, mostly linear statistical models, the reconstructed meshes only capture the coarse geometry shape while  subtle details are missing.
For better generalization to unconstrained conditions, \cite{t19fml, tran2019towards} jointly learn a 3D prior and reconstruct 3D faces from images.
Although monocular reconstruction methods can provide visually appealing 3D face reconstructions, their accuracy and quality is not suitable for applications which require metrically accurate geometry.
Recently published work indicates that existing state-of-the-art monocular 3D face reconstructions are metrically worse or only marginally better compared to a static model mean face, when compared to ground truth 3D scans~\cite{RingNet:CVPR:2019}.
This comes at little surprise as inferring 3D geometry from a single image is an ill-posed problem due to the inherent ambiguity of focal length, scale and shape~\cite{bas2017does} as under perspective projection different shapes result in the same image for different object-camera distances.
Our method instead leverages explicit calibrated multi-view information to reconstruct metrically accurate 3D geometry.

\begin{figure*}[ht]
\begin{center}
	\includegraphics[width=0.8\textwidth, trim={0cm, 2.0cm, 0cm, 0cm}]{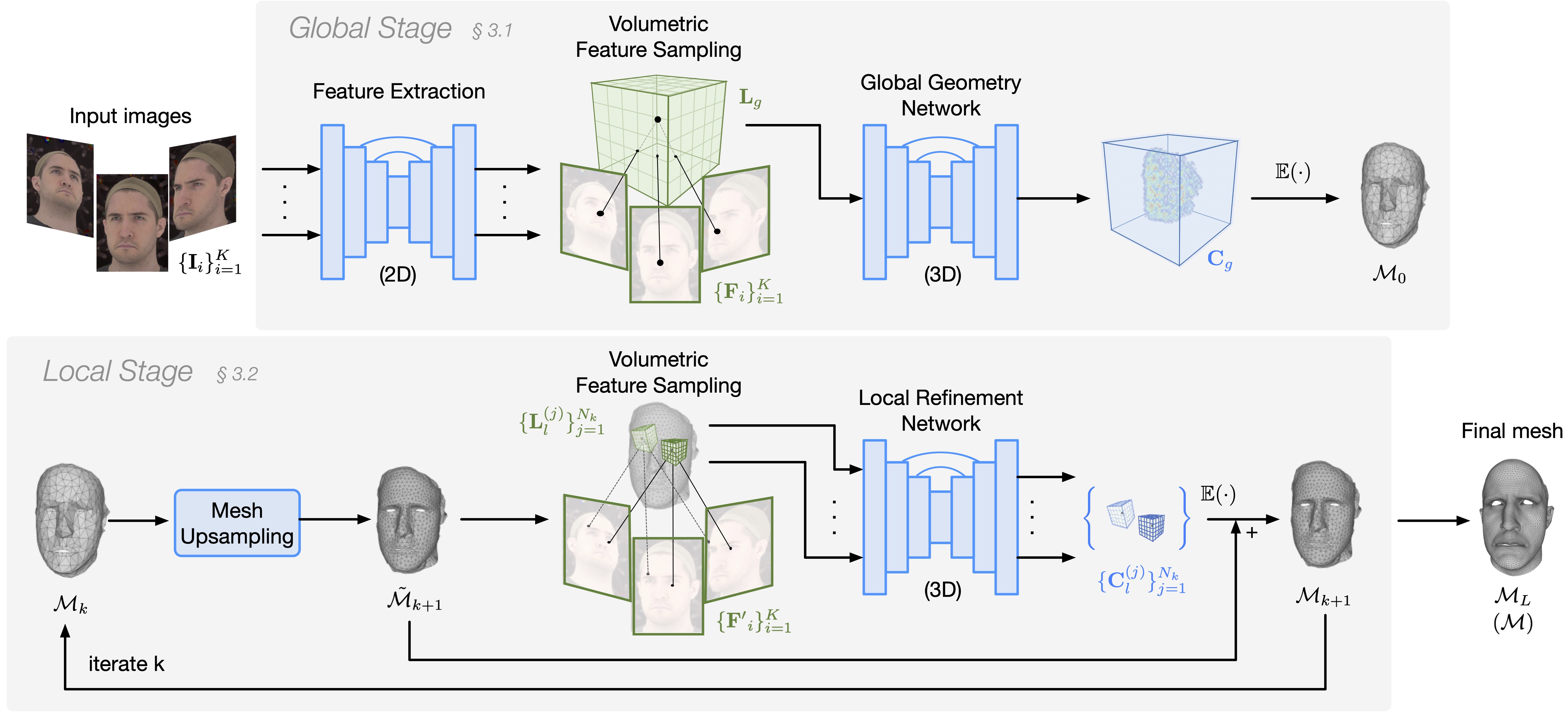}
\end{center}
	\caption{Overview of the progressive mesh generation network. 
	}
	  \label{fig:pipeline}
	  \vspace{-10pt}
\end{figure*}

\section{Multi-View Face Inference}
\label{sec:method_geometry}
As shown in Fig.~\ref{fig:pipeline_system}, given images $\{\mathbf{I}_i\}_{i=1}^{K}$ in $K$ views with known camera calibration $\{\mathbf{P}_i\}_{i=1}^{K}$, together denoted as $\mathcal{I} = \{\mathbf{I}_i, \mathbf{P}_i\}_{i=1}^{K}$, the goal of \name is two-fold:
(1) to reconstruct an accurate \textit{base mesh} in an artist-designed topology, and 
(2) to estimate pore-level geometric details and high-quality facial appearance in form of albedo and specular reflectance maps.
Formally, an output base mesh $\mathcal{M}$ contains a list of vertices $\mathbf{V} \in \mathbb{R}^{N \times 3}$ and a fixed triangulation $\mathbf{T}$.
The base meshes are required to  
(1) tightly fit the face surfaces,  
(2) share a common artist-designed mesh topology, where each vertex encodes the same semantic interpretation across all meshes, and
(3) have a sufficient triangle or quad density (with $N > 10^4$ number of vertices).

The key to dense mesh prediction is a coarse-to-fine network architecture, as shown in Fig.~\ref{fig:pipeline}.
The desired semantic mesh correspondence is naturally embedded in the hierarchical architecture.
Based on that, the geometry is inferred by the following two stages:
(1) a coarse mesh prediction $\mathcal{M}_{0}$, by the global stage $\mathbf{V}_{0} = \mathcal{F}_{g}(\mathcal{I})$;
and (2) iteratively upsampling and refining into the denser meshes $\{\mathcal{M}_{1}, \mathcal{M}_{2}, ..., \mathcal{M}_{L}\}$, by the local stage $\mathbf{V}_{k+1} = \mathcal{F}_{l}(\mathcal{I}, \mathbf{V}_{k})$.
$\mathcal{M}_{L}$ is the final prediction of base mesh $\mathcal{M}$.

Conceptually, the global stage mimics a learning-based MVS, while the local stage provides ``updates'' as if in an iterative mesh registration.
In contrast to the two traditional methods, our two steps share  consistent correspondence in a fixed topology and use volumetric features for geometry inference and surface refinement.

\subsection{Global Geometry Stage}

\qheading{Volumetric Feature Sampling.}
In order to extract salient features to predict surface points in correspondence, we deploy a shared U-Net convolutional network to extract local 2D feature maps $\mathbf{F}_i$ for each input image $\mathbf{I}_i$.
We sample volumetric features $\mathbf{L}$ by bilinearly sampling and fusing image features at projected coordinates in all images for each local point $\mathbf{v} \in \mathbb{R}^3$ in the 3D grid $\mathcal{G}$:
\begin{equation}
    \label{eq:volume_sam}
    \mathbf{L}(\mathbf{v})
        = \sigma(
        \{
            \mathbf{F}_i(
                \Pi(
                    \mathbf{v}, \mathbf{P}_i
                )
            )
        \}_{i=1}^{K}
    ),
\end{equation}
where $\Pi(\cdot)$ is the perspective projection function and $\sigma(\cdot)$ is a view-wise fusion function, for which common choices can be max, mean or standard deviation.
The 3D grid $\mathcal{G}$ is a set of points on a regular 3D grid, which can be defined at arbitrary locations with arbitrary shapes.
Here we choose cube grids, as shown in green cubes in Fig.~\ref{fig:pipeline} to feed into 3D convolution networks.

\qheading{Global Geometry Network.}
To enable the vertex flexibility, we design the network to predict vertex location free of the constraint of 3DMMs.
To encourage better generalization, we design a volumetric network architecture to learn the probabilistic distribution instead of the absolute location for each vertex.
We define a canonical global grid $\mathcal{G}_g$ that covers the whole captured volume for subject heads.
We apply the volumetric feature sampling (Eq.~\ref{eq:volume_sam}) on the global grid $\mathcal{G}_g$ to obtain the global volumetric feature $\mathbf{L}_g$, similar to \cite{im2019deep, iskakov2019learnable}.
We deploy the global geometry network $\Phi_{g}$, a 3D convolutional network with skip connections, to predict a probability volume $\mathbf{C}_g = \Phi_{g}(\mathbf{L}_g)$, in which each channel encodes the probability distribution for the location of a corresponding vertex in the initial mesh $\mathcal{M}_0$.
The vertex locations are extracted by a per-channel soft-argmax operation, $\mathbf{V}_{0} = \mathop{\mathbb{E}}(\mathbf{C}_g)$, similar to that in \cite{iskakov2019learnable}.

\subsection{Local Geometry Stage}

\begin{figure}[ht]
	\centering
    \includegraphics[width=0.95\linewidth]{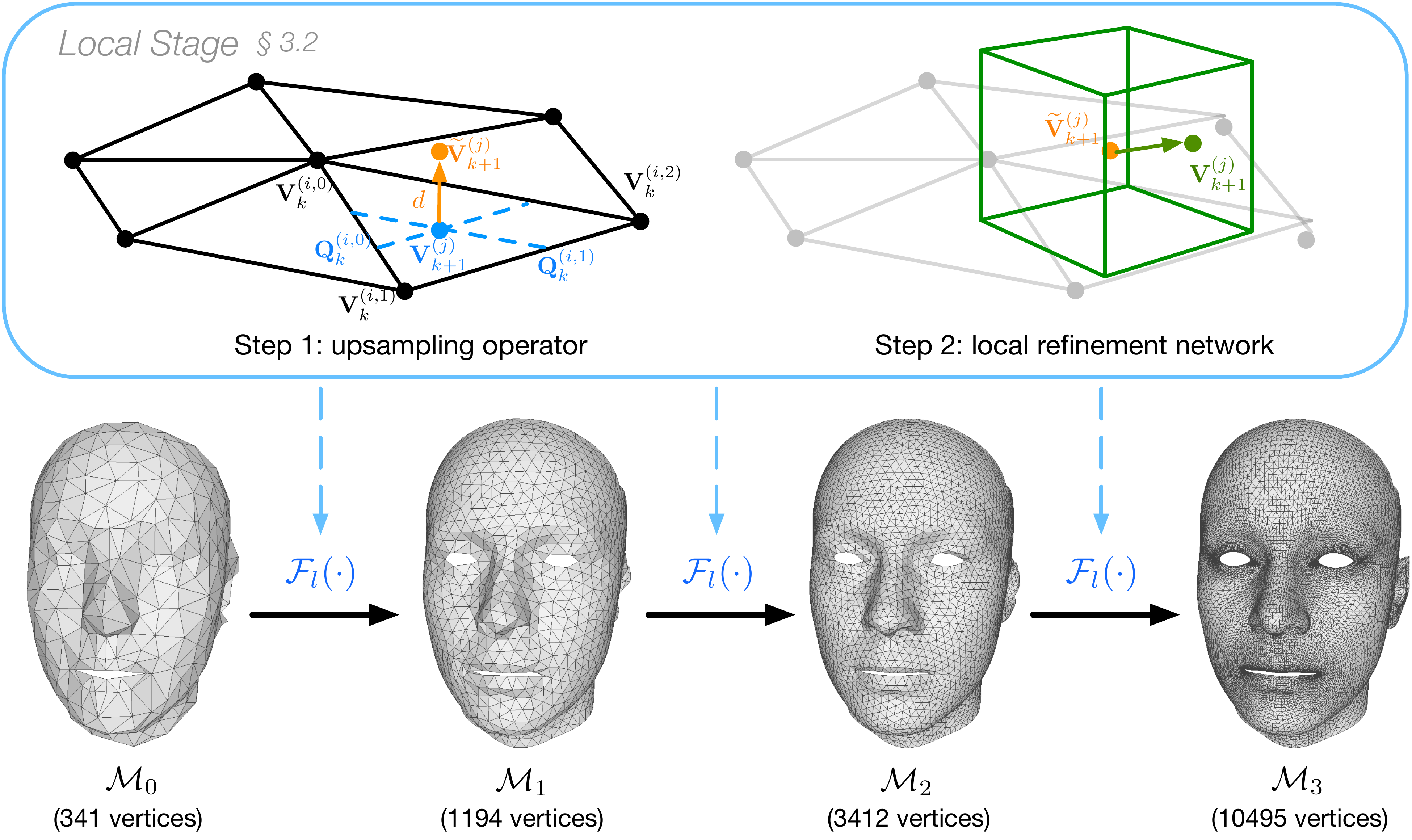}
	\caption{The iterative upsampling and refinement process in the local geometry stage.
	}
	  \label{fig:overview_upsample}
\end{figure}

Based on the coarse mesh $\mathcal{M}_{0}$ obtained from the global stage, the local stage progressively produces meshes in higher resolution and with finer details, $\{\mathcal{M}_{k}\}_{k=1}^{L}$.
At each level $k$, this process is done in two steps, as shown in Fig.~\ref{fig:overview_upsample}:
(1) a fixed and differentiable upsampling operator to provide a reliable initialization for upsampled meshes,
and (2) a local refinement network to further improve the surface details based on the input images.

\qheading{Upsampling Operator.}
Ranjan et al.~\cite{CoMA2018} propose a mesh upsampling technique based on the barycentric embedding of vertices in the lower-resolution mesh version. 
Directly using this upsampling scheme results in unsmooth  artifacts, as the barycentric embedding contrains the upsampled vertices to lie in the surface of the lower-resolution mesh.
Instead, we use additional normal displacement weights as shown in step 1 of Fig.~\ref{fig:overview_upsample}.
Given a sparser mesh $\mathcal{M}_{k} = (\mathbf{V}_k, \mathbf{T}_k)$ and its per-vertex normal vectors $\mathbf{N}_k$, we upsample the mesh by
\begin{equation} \label{eq:upsample_op}
    \widetilde{\mathbf{V}}_{k+1} =  
        \mathbf{Q}_{k} \mathbf{V}_{k} +
        \mathbf{D}_{k} \mathbf{N}_{k}
,
\end{equation}
where $\mathbf{Q}_{k} \in \mathbb{R}^{N_{k+1} \times N_{k}}$ is the barycentric weight matrix as in \cite{CoMA2018} and $\mathbf{D}_{k-1}  \in \mathbb{R}^{N_{k+1} \times N_{k}}$ is the additional coefficient matrix that apply displacement vectors along normal directions. 
The normal displacements encode additional surface details that allow vertices to be outside of the input surface.

For a hierarchy with $L$ levels, we first downsample the full-resolution template mesh $\mathcal{T} = (\mathbf{V}, \mathbf{T}) :  = \mathcal{T}_{L}$ by isotropic remeshing and non-rigid registration, into a series of meshes with decreasing resolution while still preserving geometry and topology of original mesh: $ \{\mathcal{T}_{L-1}, \mathcal{T}_{L-2}, ..., \mathcal{T}_{0} \} $.
Next, we embed the vertices at higher resolution in the surface at lower resolution meshes by barycentric coordinates $\mathbf{Q}_{k}$ as in \cite{CoMA2018}.
We then project the remaining residual vectors onto the normal direction and obtain $\mathbf{D}_{k}$.

\qheading{Local Refinement Network.} 
Around each vertex (indexed with $j$) of the upsampled mesh $\widetilde{\mathbf{V}}_{k+1}^{(j)}$, we define a smaller grid than $\mathcal{G}_g$ in the global stage in the local neighborhood $\mathcal{G}_{l}^{(j)}$.
We sample local volumetric features $\mathbf{L}_{l}^{(j)}$ by Eq.~\ref{eq:volume_sam}.
For each local feature volume, we apply the local refinement network $\Phi_{l}$, a 3D convolutional network with skip connections, to predict per-vertex probability volume $\mathbf{C}_{l}^{(j)} = \Phi_{l}(\mathbf{L}_{l}^{(j)})$. 
Then we compute the corrective vector by the expectation operator, $\delta \mathbf{V}_{k+1}^{(j)} = \mathop{\mathbb{E}}(\mathbf{C}_{l}^{(j)})$.
This process is applied to all vertices independently, and therefore can be parallelized in batches.
Finally the upsampled and refined mesh vertices are
\begin{equation} \label{eq:local_stage}
    \mathbf{V}_{k+1} = 
    \widetilde{\mathbf{V}}_{k+1} +
    \delta \mathbf{V}_{k+1}.
\end{equation}
 Given $\mathcal{M}_{0}$, we iteratively apply the local stage at all levels until we reach the highest resolution and obtain $\mathcal{M}_{L}$.

The volumetric feature sampling and the upsampling operator, along with the networks are fully differentiable, enabling the progressive geometry network end-to-end trainable from input images to dense registered meshes.

\subsection{Appearance and Detail Capture}
Skin detail and appearance maps are commonly used in photo-realistic rendering, which is often difficult to estimate without special capture hardware, such as the Light Stage capture system \cite{debevec2000acquiring}.
We propose a simple yet effective architecture to estimate high-resolution detail and appearance maps, potentially without the dependency on special appearance capture systems.

\qheading{Albedo Maps Generation.}
The base meshes are reconstructed for a smaller head region.
We augment the base meshes by additional fitting for the back of the head using Laplacian deformation \cite{sorkine2004laplacian}. 
We then perform the standard texturing given the completed mesh and multi-view images and obtain the albedo reflectance map on the UV domain.
Furthermore, by applying the texturing process and sample vertex locations instead of RGB colors, we obtain another map on the UV domain, that we call the geometry map.

\qheading{Detail Maps Synthesis.}
To further augment the representation, we adopt an image-to-image translation strategy to infer finer-level details. 
Using a network similar to \cite{wang2018pix2pixHD}, our synthesis network infers specular reflectances and displacements given both albedo and geometry map. 
We then upscale all the texture maps to 4K resolution by using the super resolution strategy of \cite{wang2018esrgan}. 
We can obtain the detailed mesh in high-resolution by applying the displacement maps on the base mesh, as shown in Fig.~\ref{fig:pipeline_system}.
The reconstructed skin detail and appearance maps are directly usable for standard graphics pipelines for photo-realistic rendering.

\section{Experiments}
\label{sec:result}

\paragraph{Datasets.}
\name is trained and evaluated on datasets captured with a Light Stage system~\cite{Ghosh_2011_SIGGRAPH, ma2007rapid}, with 3D scans from MVS, ground truth base meshes from a traditional mesh registration pipeline~\cite{RuilongLi2020}, and ground truth skin attributes from the traditional light stage pipeline \cite{debevec2000acquiring}. 
In particular, we correct the ground truth base meshes with optical flow and manual work of a professional artist, to ensure high quality and high accuracy of registration.
The dataset contains 64 subjects (45 for train and 19 for test), covering a wide diversities in gender, age and ethnicity.
Each set of capture contains a neutral face and 26 expressions, including some extreme face deformations (e.g. mouth widely open), asymmetrical motions (jaw to left/right) and subtle expressions (e.g. concave cheek or eye motions).

\begin{figure*}[ht]
	\centering
	\includegraphics[width=1.0\linewidth]{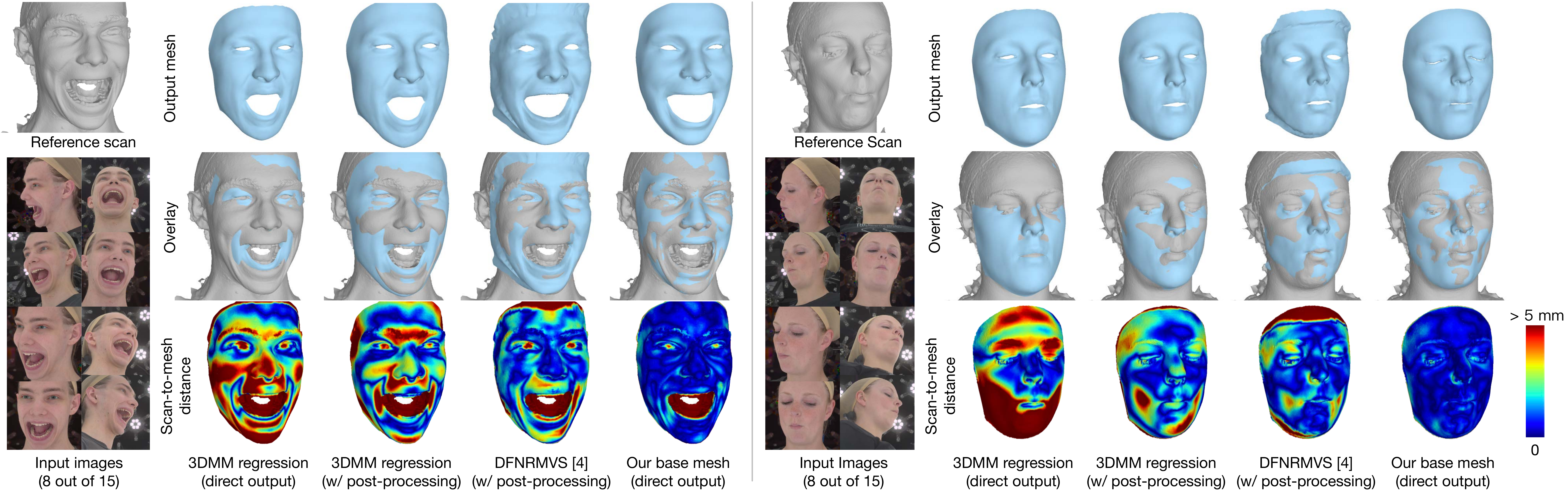}
	\caption{
	Qualitative comparison on geometric accuracy with the existing methods.
	The scan-to-mesh distance is visualized in heatmap (red means $>5$ mm).
	Note that 3DMM and DFNRMVS \cite{bai2020dfnrmvs} need rigid ICP as post-processing.
	Our outputs require no post-processing, while outperforming the existing learning-based method in geometry accuracy.
	}
	  \label{fig:expr_geo_visual}
	  \vspace{-5pt}
\end{figure*}

\paragraph{Implementation Details.}
For the progressive mesh generation network, our feature extraction network adopts a pre-trained UNet \cite{ronneberger2015u} with ResNet34 \cite{he2016deep} as its backbone, which predicts feature maps of same resolution of input image with 8 channels. %
The volumetric features of the global stage are sampled from a $32^3$ grid with grid size of 10 millimeters, the local stage uses a $8^3$ grid with a grid size of 2.5 millimeters.
We randomly rotate the grids for the volumetric feature sampling as data augmentation during training.
The mesh hierarchy with $L=3$ contains meshes with 341, 1194, 3412 and 10495 vertices.
Both the global geometry network and local refinement network use a similar architecture as the V2V network in \cite{iskakov2019learnable}. 
Both stages are trained separately. 
The global stage trains for 400K iterations with a $l_2$ loss $\norm{\mathbf{V}_{0} - \bar{\mathbf{V}}_{0}}_{2}^{2}$, the local stage trains for 150K iterations with a $l_2$ loss combined across mesh hierarchy levels with equal weights,
$
    \sum_{k=0}^{L} \norm{
        \mathbf{V}_{k} - \bar{\mathbf{V}}_{k}
    }_{2}^{2}
$, 
where $\bar{\mathbf{V}}_{k}$ is the ground truth base mesh vertices for the predicted $\mathbf{V}_{k}$ at level $k$.
We train the progressive mesh generation network using Adam optimizer with a learning rate of $1e-4$ and batch size of 2 on a single NVIDIA V100 GPU.
For the detail maps synthesis, we adopt the synthesis network from \cite{wang2018pix2pixHD} and the super-resolution network from ESRGAN \cite{wang2018esrgan}. For more details, see the \textit{Sup. Mat.}

\subsection{Results}

\qheading{Baselines.}
We evaluate the performance of our base mesh prediction and compare to the following existing methods:
(1) \textbf{Traditional MVS and Registration:} we run commercial photogrammetry software AliceVision \cite{Alicevision}, followed by non-rigid ICP surface registration. 
(2) \textbf{3DMM Regression:} we adopt a network architecture similar to  \cite{Tewari2018, tewari2018self, wu2019mvf} for a multi-view setting.
(3) \textbf{DFNRMVS \cite{bai2020dfnrmvs}:} a method that learns an adaptive model space and on-the-fly iterative refinements on top of 3DMM regression.

We argue that the two-step methods of MVS and registration is susceptible to MVS errors and requires manual tweaking optimization parameters for different inputs, which makes it not robust. %
Our method shows robustness and generalizability for challenging cases, outperforms existing learning-based method and achieves the state-of-the-art geometry and correspondence quality.
Our method has efficient run-time. 
We show various ablation studies to validate the effectiveness of our design. 
We will provide more comparison and results in the \textit{Sup. Mat.}

\begin{figure}[t]
    \centering
    \includegraphics[width=1\linewidth]{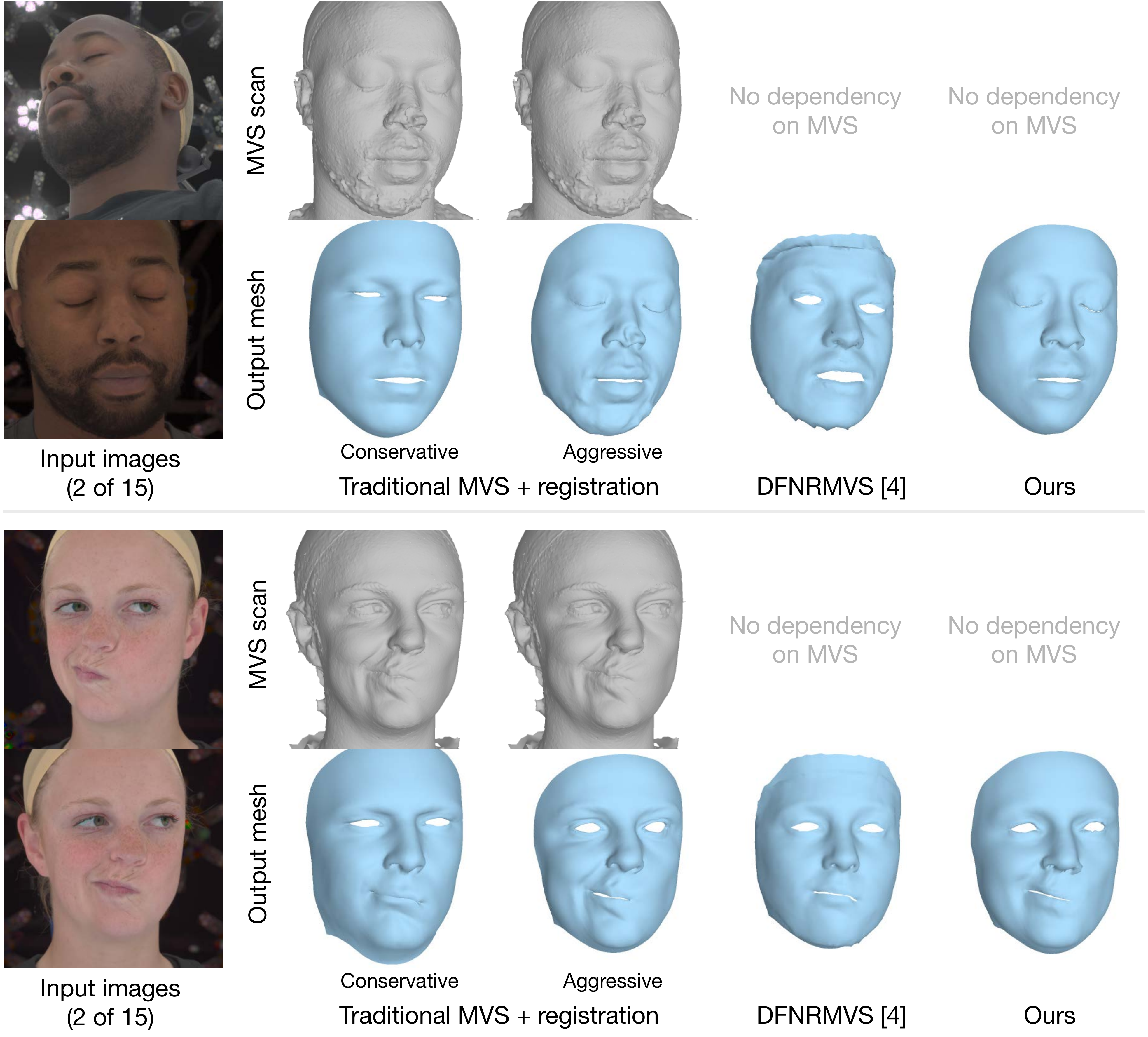}
    \caption{Evaluation on method robustness.
    }
    \label{fig:face_robust}
    \vspace{-0.2cm}
\end{figure}

\qheading{Robustness.}
Fig.~\ref{fig:face_robust} show the results from various methods given challenging inputs.
Note that when the nose of the subject (top case) is specular reflective (due to oily skin) or has facial hair, the traditional MVS fails to reconstruct the true surface, producing artifacts that affect the subsequent surface registration step.
With conservative optimization parameters (e.g. strong reliance on 3DMM), the result is more robust.
However with the same parameters, it affects the flexibility for fitting detailed shape and motion for other input cases (e.g. bottom case).
Furthermore, the extreme and asymmetrical motion is challenging for fitting only within the morphable model.
This case requires ``aggressive'' fitting, in which less regularizations are applied.
Therefore we point out this dilemma of general parameters in the traditional MVS and registration affects of automation and requires much manual work for high-quality results.
The learning based method DFNRMVS \cite{bai2020dfnrmvs} shows potential for robustness and generalizability.
However, they cannot output meshes in accurate shape and expressions.
On the contrary, our model shows superior performances in predicting a reliable mesh, given such challenging inputs.
Note that the details, such as closed eyelids and asymmetrical mouth motion are faithfully captured.

\qheading{Geometric Accuracy.}
Fig.~\ref{fig:expr_geo_visual} shows the inferred meshes given images from 15 views, along with error visualizations with the reference scans.
The 3DMM regression method cannot fit extreme or subtle expressions (wide mouth open, concave cheek and eye shut).
The adaptive space and the online refinement improve DFNRMVS \cite{bai2020dfnrmvs} for a better fitting, but it still lacks the accuracy to cover the geometric details.
Our method is capable of predicting base meshes that closely fit the ground truth surfaces.
The results recover identities for the subjects and captures challenging expressions such as extreme mouth opening or subtle non-linearity of small muscles movement (concave cheek) which cannot be modeled by linear 3DMMs. 
The overlay and error visualizations indicate that our reconstruction fits the ground truth scan closely with fitting errors significantly below 5 millimeters.
Due to not being able to utilize true projection parameters, the results of 3DMM regression and DFNRMVS \cite{bai2020dfnrmvs} lack accuracy in absolute coordinate and need a Procrustes analysis (scale and rigid pose) as post-processing for further fitting to the target.
In contrast, our method outperforms these methods \textit{without} post-processing.

As a quantitative evaluation, we measure the distribution of scan-to-mesh distances.
78.3\% of vertices by our methods have scan-to-mesh distance lower than 1 mm.
This result outperforms the 3DMM regression which have 27.0\% and 33.1\% (without and with post-processing). 
The median scan-to-mesh distance for our results is 0.584 mm, achieving sub-millimeter performance.
We show cumulative scan-to-mesh distance curves in the \textit{Sup. Mat.}

\begin{figure}[t]
    \centering
    \includegraphics[width=1.0\linewidth]{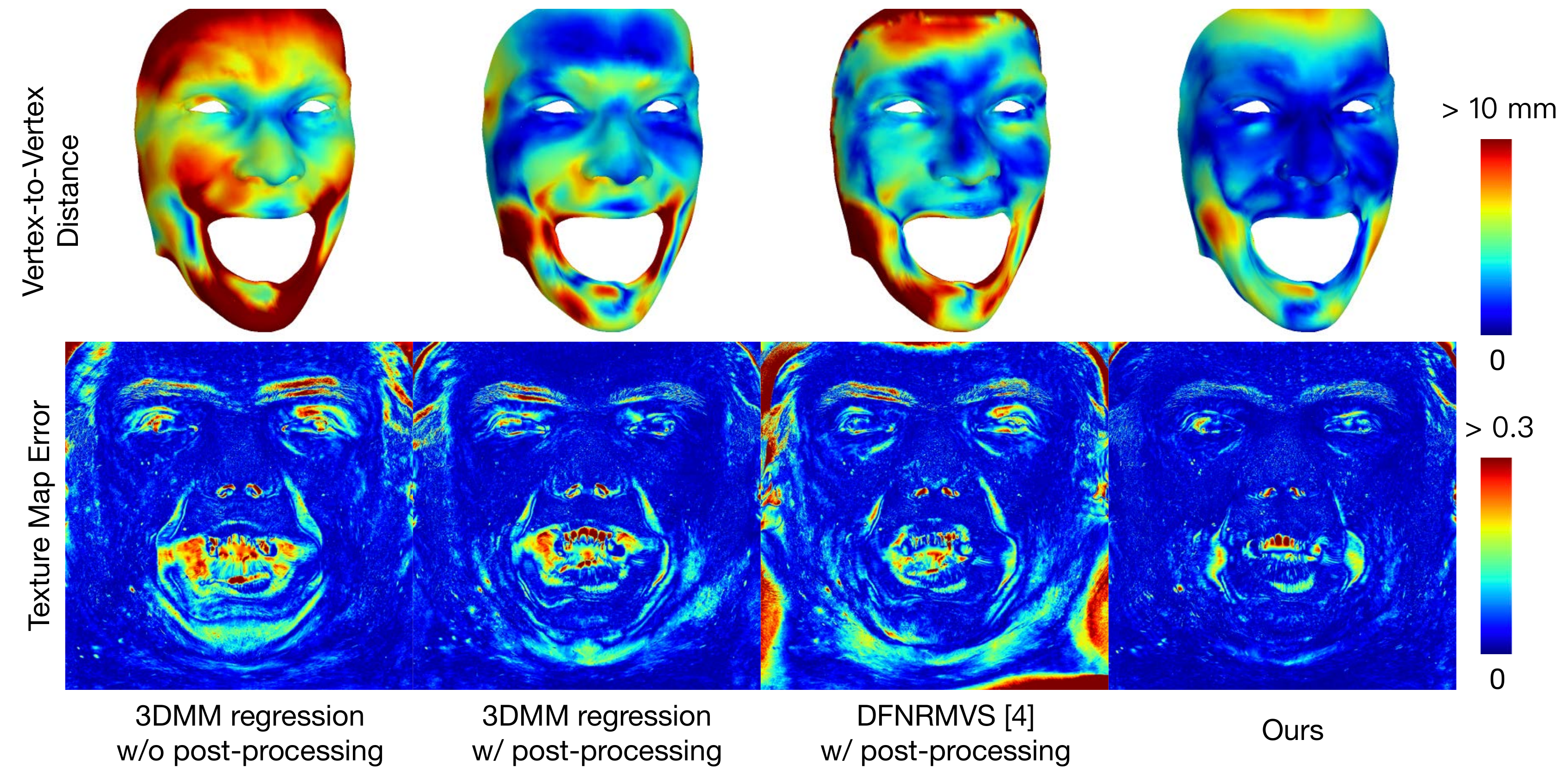}
    \caption{Visualization on correspondence accuracy}
    \label{fig:expr_corresp_visual}
\end{figure}
\begin{figure}[t]
	\centering
	\includegraphics[width=0.9\linewidth]{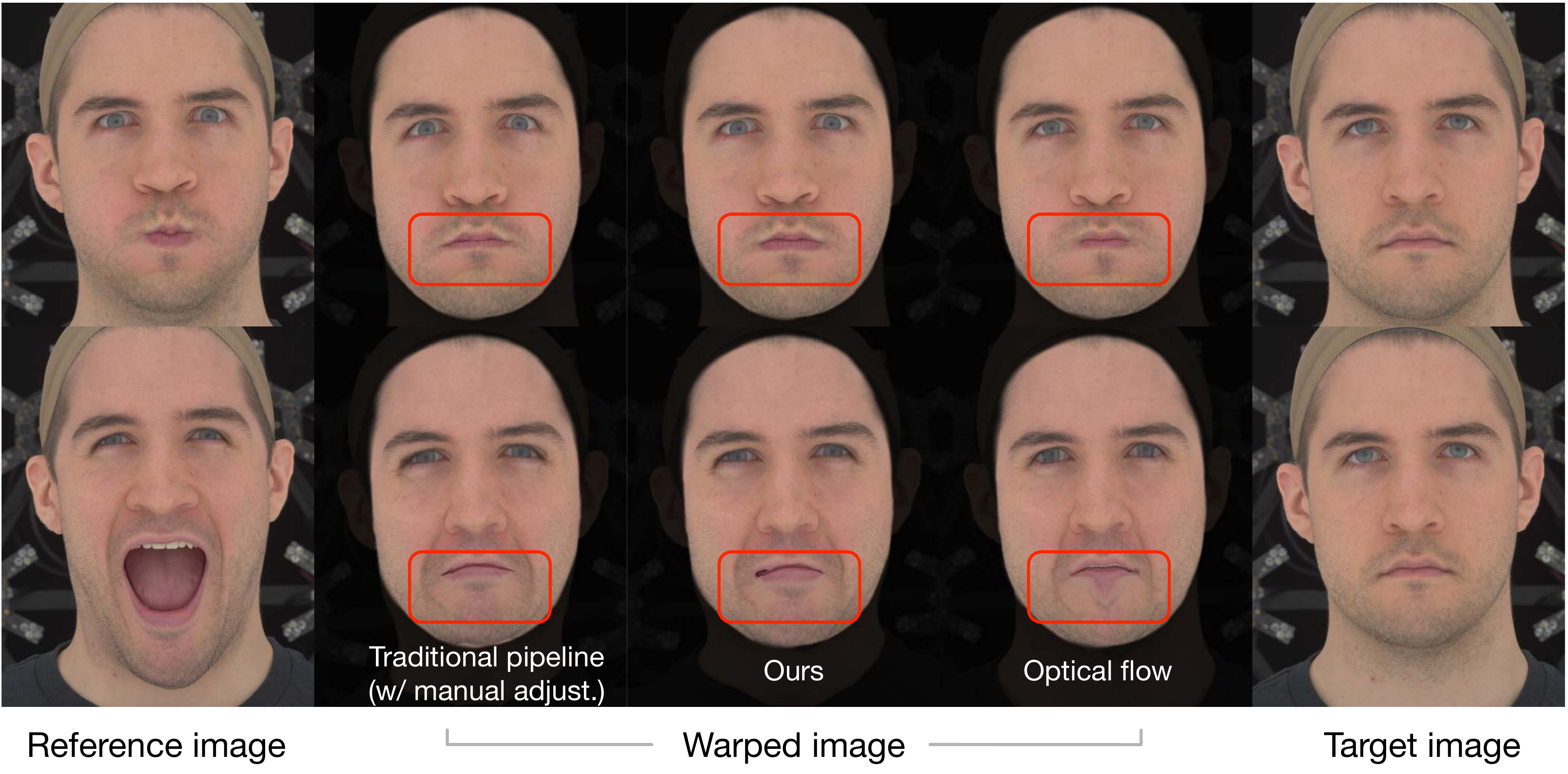}
	\caption{Qualitative evaluation on correspondence compared to optical flow.}
	  \label{fig:optical_flow}
\end{figure}

\qheading{Correspondence Accuracy.}
We provide quantitative measure for correspondence accuracy for generated base meshes, by comparing them to the ground truth aligned meshes (artist-generated in the same topology), and compute the vertex-to-vertex (v2v) distances on a test set.
The 3DMM regression method achieves a median v2v distance of 3.66 mm / 2.88 mm (w/o and w/ post-processing).
Our method achieves 1.97 mm outperforming the existing method.
The v2v distances are also visualized on the ground truth mesh in Fig.~\ref{fig:expr_corresp_visual}.
We additionally evaluate our aligned meshes by the median errors to the ground truth 3D landmarks.
Our method achieves 2.02 mm, while the 3DMM regression method achieves 3.92 mm / 3.21 mm (w/o and w/ post-processing).
We provide more quantitative evaluations in the \textit{Sup. Mat.} 

We compute the photometric errors between the texture map of the output meshes and the one of the ground truth meshes.
Lower photometric errors indicate the UV textures match the pre-designed UV parametrization (i.e. better correspondence).
Our method has significantly lower errors, especially in the eyebrow region, around the jaw and for wrinkles around eyes and nose.
Note that the 3DMM regression method without post-processing performs worse, while our method requires no post-processing.

In Fig.~\ref{fig:optical_flow}, we further evaluate the correspondence quality by projecting it onto 2D images and warping the reference image (extreme expression) back to target image (neutral expression).
The ideal warping outputs would be as close to the target image as possible, except for shades as in wrinkles.
We compare the performance with traditional pipeline of MVS and registration (with manual adjustment) and the traditional optical flow method. 
Our method recovers better 2D correspondence than optical flow, which relies on local matching which tends to fail when occlusion and large motion, as shown in Fig.~\ref{fig:optical_flow} (see lip regions).
Further optical flow takes 30 seconds on image resolution 1366 × 1003, compared within 1 second based on our base meshes.
The traditional method achieves good results, but at a cost of 3 orders of magnitude longer of processing time and possibly manual adjustment.

\begin{table}[t]
\centering
\addtolength{\tabcolsep}{4.5pt}
\begin{tabular}{l|c|c}
\toprule
Methods & Time  & Automatic \\ 
\midrule
Traditional pipeline & 600+  & {\color{red}\xmark}  \\
DFNRMVS \cite{bai2020dfnrmvs} & 4.5 & {\color{green}\cmark} \\
\name (base mesh) & 0.385 & {\color{green}\cmark} \\
\bottomrule
\end{tabular}
\setlength{\belowcaptionskip}{-10pt}
\caption{Comparison on run time on base mesh, given images from 15 views and measured in seconds.
}
\label{tab:speed}
\end{table}

\qheading{Inference Speed.}
The traditional pipeline takes more than 10 minutes and potentially more time for manual adjustments.
DFNRMVS \cite{bai2020dfnrmvs} infers faces without tuning at test-time but is still slower at 4.5 seconds due to its online optimization step and heavy computation on the dense photometric term.
Our global and local stage takes 0.081 seconds and 0.304 seconds respectively. 
As shown in Table~\ref{tab:speed}, our method produces a high-quality registered base mesh in 0.385 seconds, and achieves sub-second performance, while being fully automatic without manual tweaking.

\begin{figure}[t]
	\centering
	\includegraphics[width=1.0\linewidth]{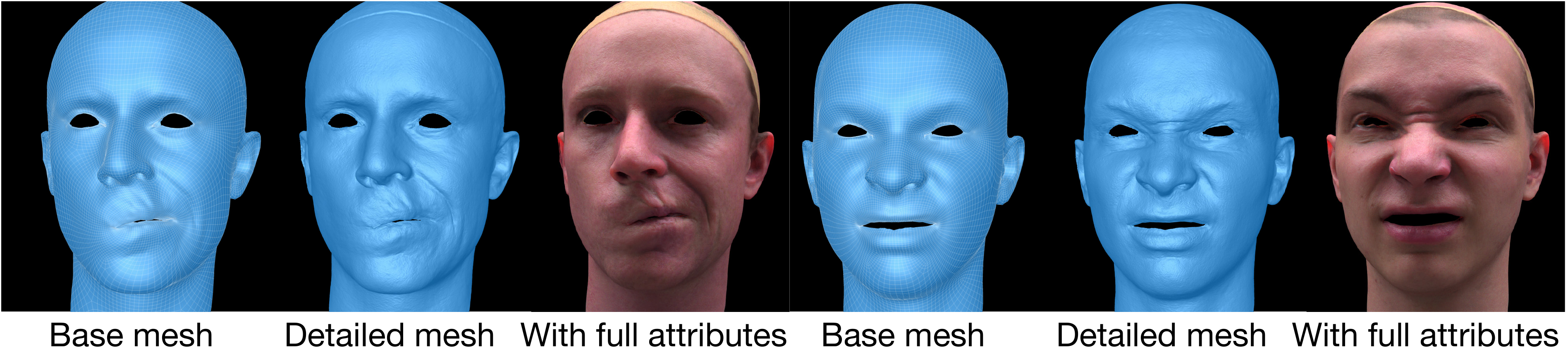}
	\caption{Based on our reliable base meshes, our appearance and detail capture network predicts realistic face skin details and attributes, without special hardware such as Light Stage at test-time, enabling photo-realistic rendering.}
	  \label{fig:face_details}
	  \vspace{-5pt}
\end{figure}

\qheading{Appearance Capture.}
In Fig.~\ref{fig:teaser} and Fig.~\ref{fig:face_details}, we show rendering results with the inferred displacement and albedo and specular maps, enabling photo-realistic renderings. 

\begin{figure}[t]
    \centering
    \includegraphics[width=1.0\linewidth]{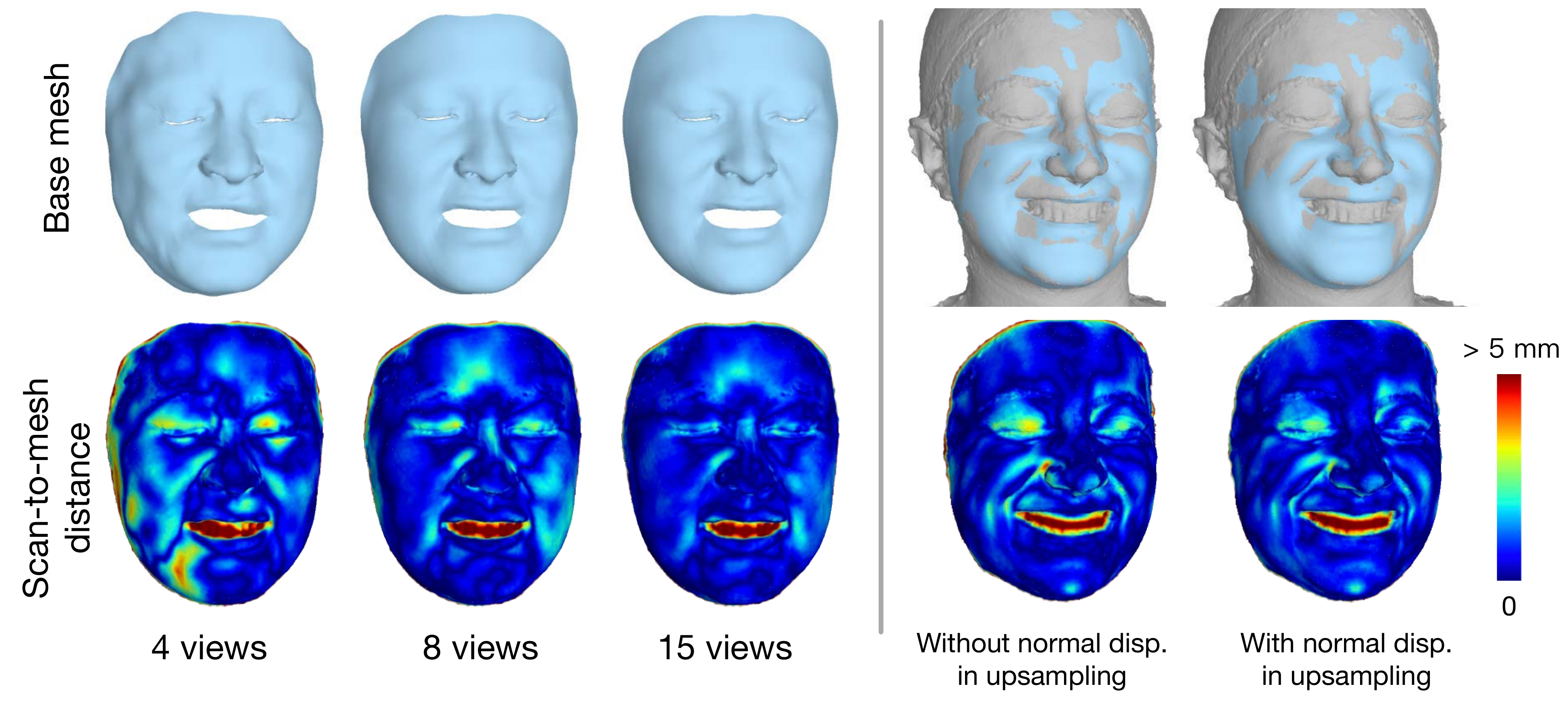}
    \caption{Ablation studies. Left: number of input camera views; Right: on normal displacement weights in mesh upsampling function.
    }
    \label{fig:expr_ablation_visual}
\end{figure}

\qheading{Ablation Studies.}
In Fig.~\ref{fig:expr_ablation_visual} (left), we evaluate the robustness of our network on various numbers of input views.
The resulting quality degrades gracefully as the views decrease. Our method produces reasonable results on views as sparse as 4, which is extremely difficult for standard MVS due to large baseline and little overlaps.
Fig.~\ref{fig:expr_ablation_visual} (right) demonstrates the normal displacement in the upsampling function contributes in capturing fine shape details.
We provide more ablation studies in the \textit{Sup. Mat.}

\begin{figure}[t]
    \centering
    \includegraphics[width=1.0\linewidth]{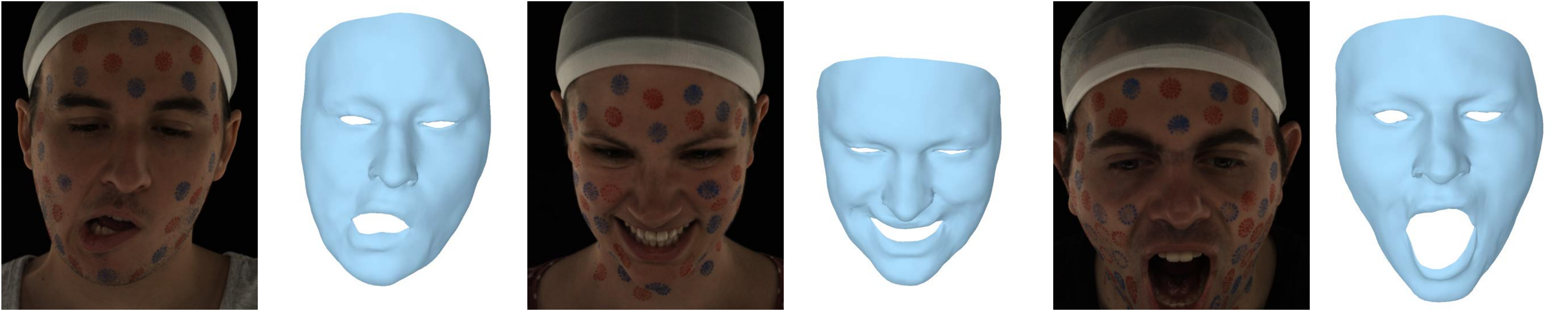}
    \caption{
    Results on CoMA \cite{CoMA2018} datasets.
    }
    \label{fig:coma_finet}
    \vspace{-0.5cm}
\end{figure}

\qheading{Generalization to New Capture Setups.}
We finetune our network on the CoMA \cite{CoMA2018} dataset, which contains a different camera setup, significantly fewer views (4) and subjects (12), different lighting conditions and special make-up patterns painted on subjects' faces.
The results in Fig.~\ref{fig:coma_finet} show that our system can in principle be applied to the different capture setups.
However, we observe some artifacts around jaws and slightly protruding eyebrow bones. %
This is potentially due to limited number of subjects and insufficient camera coverage (e.g. the 3rd image misses the jaw region).

\section{Conclusion}
\label{sec:conclusion}

We introduced a 3D face inference approach from multi-view input images that can produce high-fidelity 3D faces meshes with consistent topology using a volumetric sampling approach.
We have shown that, given multi-view inputs, implicitly learning a shape variation and deformation field can produce superior results, compared to methods that use an underlying 3DMM even if they refine the resulting inference with an optimization step.
We have demonstrated sub-millimeter surface reconstruction accuracy, and state-of-the-art correspondence performance while achieving up to 3 orders of magnitude of speed improvement over conventional techniques.
Most importantly, our approach is fully automated and eliminates the need for data clean up after MVS, or any parameter tweaking for conventional non-rigid registration techniques.
Our experiments also show that the volumetric feature sampling can aggregate effectively features across views at various scales and can also provide salient information for predicting accurate alignment without the need for any manual post-processing. 
Our next step is to extend our approach to regions beyond the skin region, including teeth, tongue, and eyes. We believe that our volumetric digitization framework can handle non-parametric facial surfaces, which could potentially eliminate the need for specialized shaders and models in conventional graphics pipelines.
Furthermore, we would like to explore video sequences, and investigate ways to ensure temporal coherency in fine-scale surface deformations.
Our model is suitable for articulated non-rigid objects such as human bodies, which motivates us to look into more general shapes and objects such as clothing and hair.

\qheading{Acknowledgement.}
We thank M. Ramos, M. He and J. Yang for the help in visualizations, and P. Prasad, Z. Li and Z. Lv for proofreading.
The research was sponsored by the Army Research Office and under Cooperative Agreement Number W911NF-20-2-0053, and sponsored by the U.S. Army Research Laboratory (ARL) under contract number W911NF-14-D-0005, the CONIX Research Center, one of six centers in JUMP, a Semiconductor Research Corporation (SRC) program sponsored by DARPA and in part by the ONR YIP grant N00014-17-S-FO14. Statements and opinions expressed and content included do not necessarily reflect the position or the policy of the Government, and no official endorsement should be inferred.

\qheading{Disclosure.}
While TB is a part-time employee of Amazon, his research was performed solely at, and funded by, MPI. 

{\small
\balance

}

\newpage
\nobalance
\appendix
\section{Appendix}

\begin{figure*}[ht]
    \centering
    \includegraphics[width=1.0\linewidth]{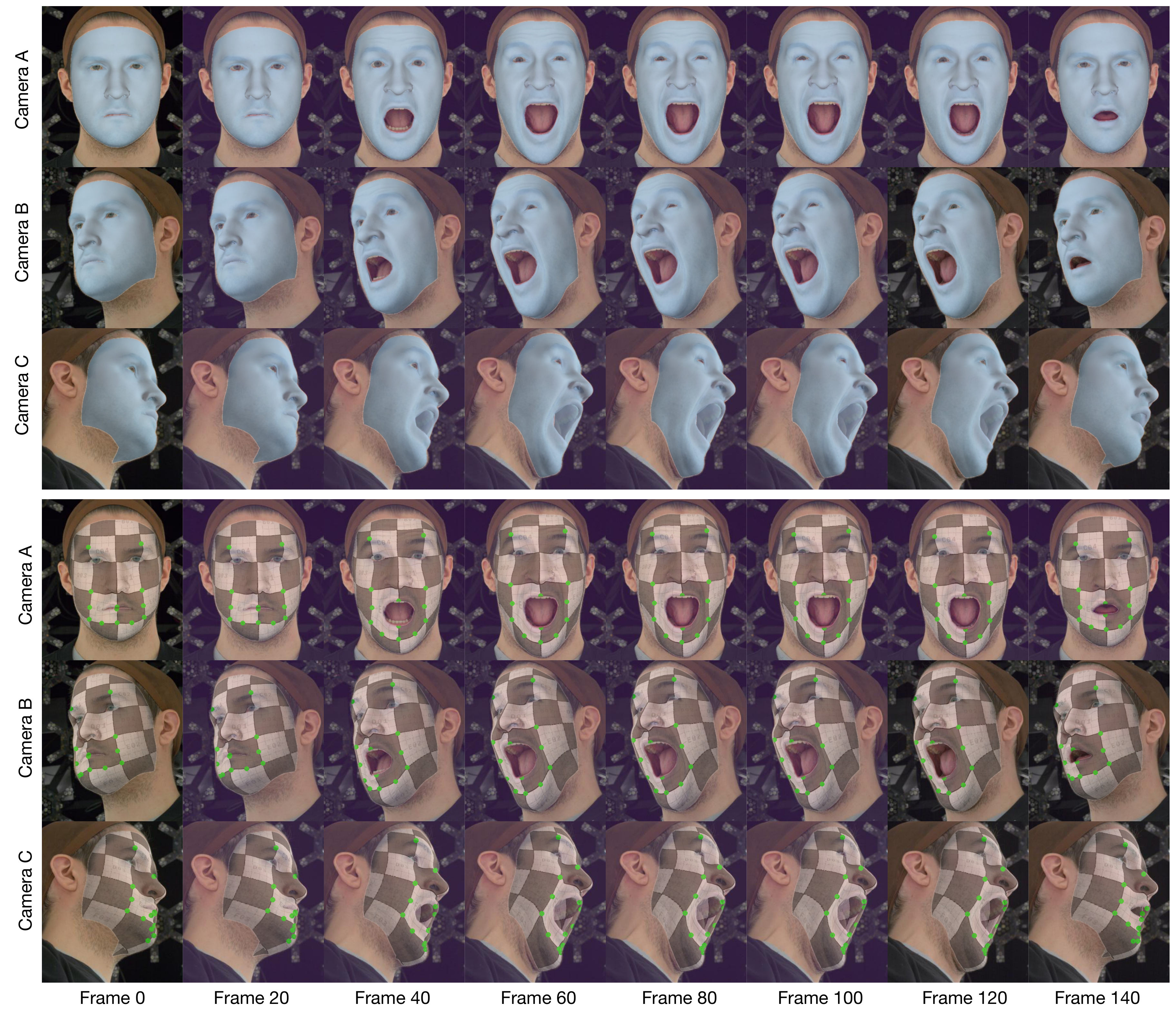}
	\caption{
	Base mesh reconstruction for a multi-view video sequence overlaid on the video frames. 
	Our method captures the facial performance well.
	The result meshes are temporally stable and accurately align with the input images.
	Visualizing with a shared checkerboard texture indicates good tracking quality.
	Please see the \textit{supplemental video} for better visualization. 
	}
	\label{fig:supple_dynamic_sequence}
\end{figure*}

\qheading{Supplemental Video.}
Please see the video on the project page: \url{https://tianyeli.github.io/tofu}.

\begin{figure}[ht]
	\centering
	\includegraphics[width=1\linewidth]{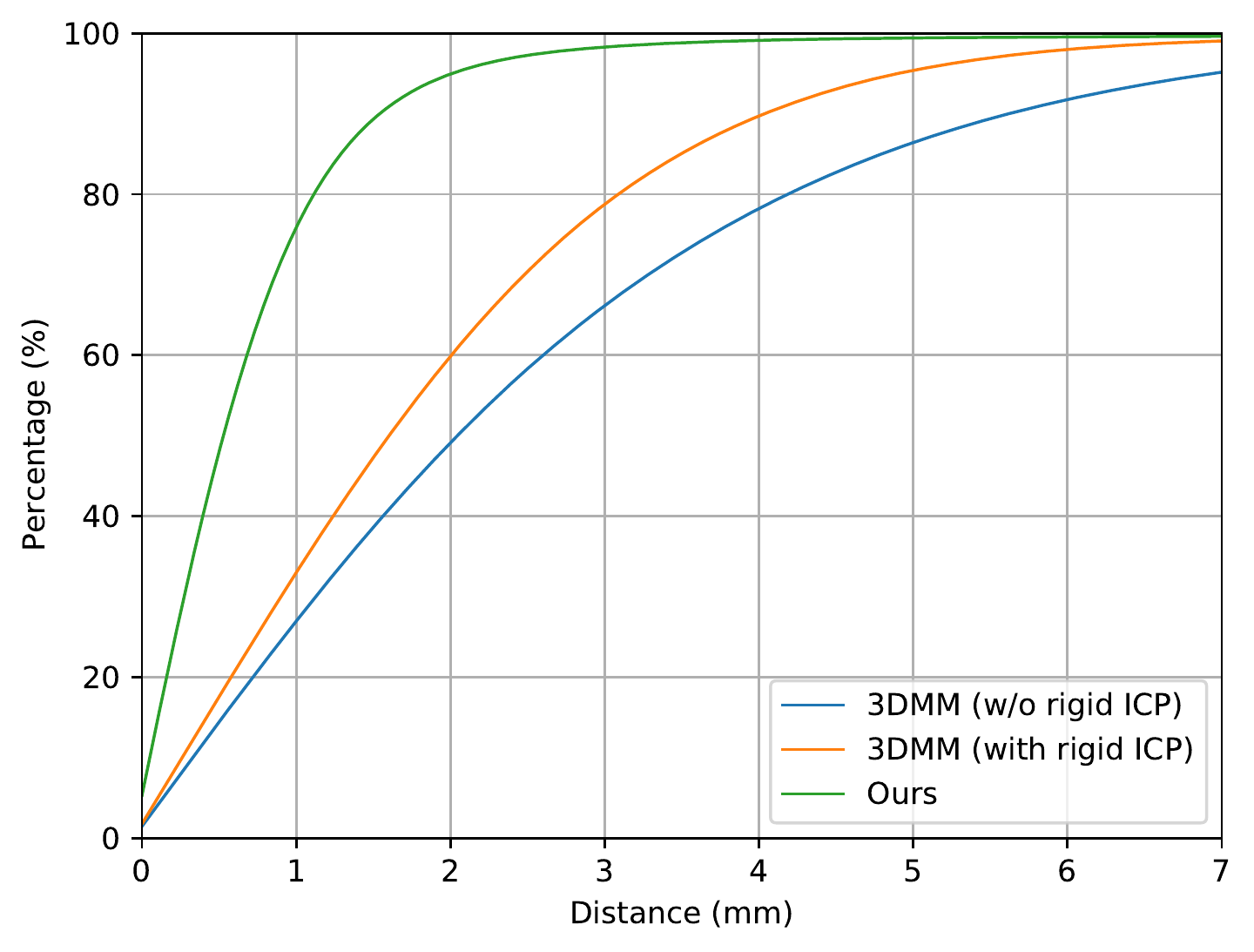}
	\caption{Quantitative evaluation by cumulative error curves for scan-to-mesh distances among learning based methods.}
	  \label{fig:supple_expr_geo_curves_methods}
\end{figure}

\qheading{Additional Quantitative Results.}
Tab.~\ref{tab:more_comparisons} provides additional quantitative comparisons to other learning based methods, namely 3DMM regression and DFNRMVS \cite{bai2020dfnrmvs}.
Fig.~\ref{fig:supple_expr_geo_curves_methods} shows the cumulative error curves for scan-to-mesh distances among the methods.
All methods are evaluated on a common held-out test set with 499 ground truth 3D scans; no data of test subjects are used during training. 
The geometric reconstruction accuracy is evaluated using scan-to-mesh distance (s2m) that measures the distance between each vertex of a ground truth scan, and the closest point in the surface of the reconstructed mesh. 
The correspondence accuracy is evaluated using a vertex-to-vertex distance (v2v) that measures the distance between each vertex of a registered ground truth mesh, and the semantically corresponding point in the reconstructed mesh. 

\begin{table}[ht]
    \centering
    \begin{tabular}{|l|c|c|c|c}
    \hline
    Methods & median s2m & median v2v \\ \hline
    3DMM Regr. & 2.104 & 3.662  \\
    3DMM Regr. (PP) & 1.659 & 2.890  \\
    DFNRMVS \cite{bai2020dfnrmvs} (PP) & 1.885 & 4.565  \\
    Our Method & \bf{0.585} & \bf{1.973} \\
    \hline
    \end{tabular}
    \caption{Comparison on geometry accuracy (median s2m), correspondence accuracy (median v2v) among the learning based methods, measured in millimeters.
    ``PP'' denotes the result after a post-processing Procrustes alignment that solves for the optimal rigid pose (i.e. 3D rotation and translation) and scale to best align the reconstructed mesh with the ground truth. 
    Note that our method requires no post-processing.
    }
    \label{tab:more_comparisons}
\end{table}

\begin{figure}[ht]
	\centering
	\includegraphics[width=0.9\linewidth]{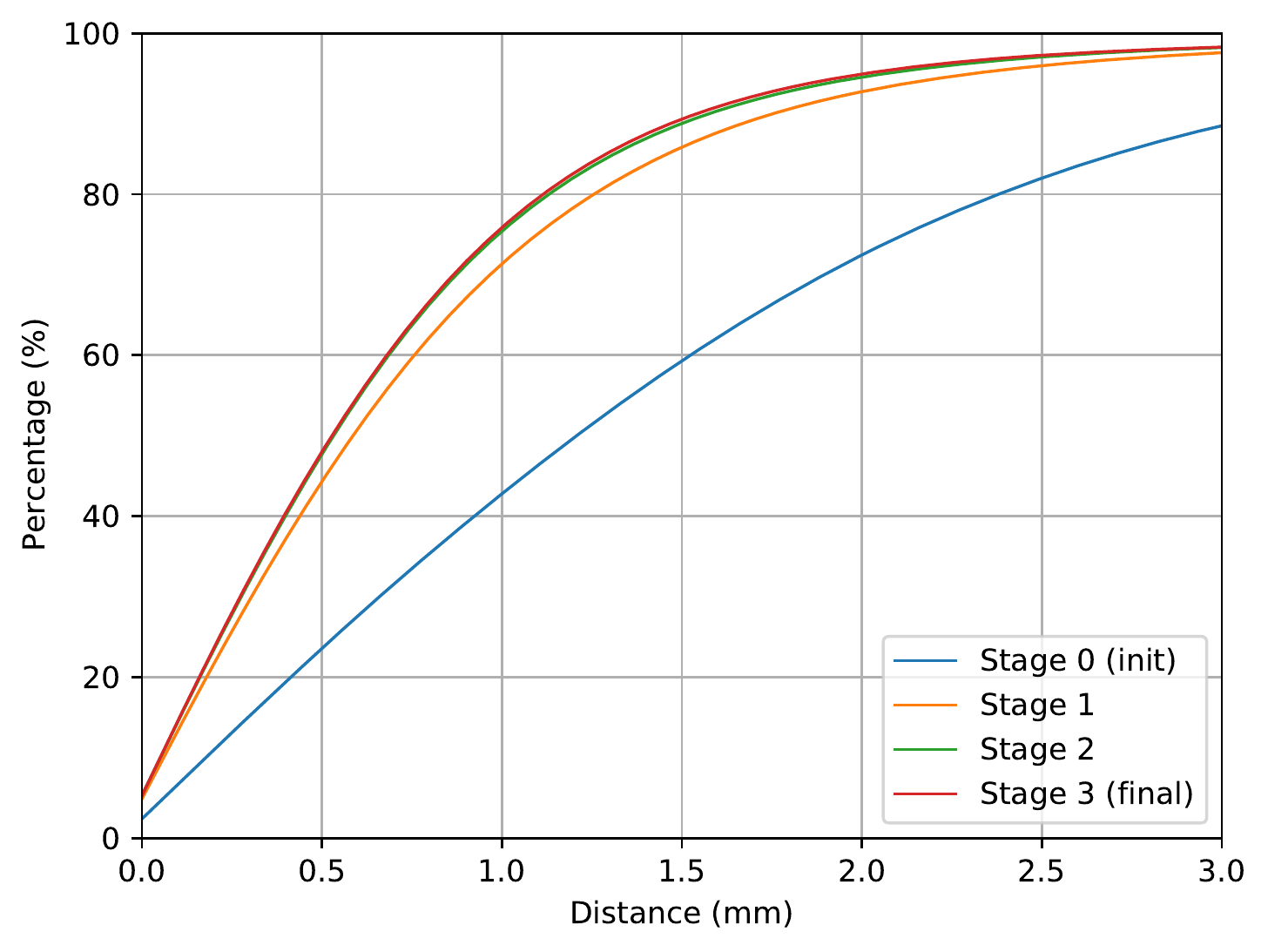}
	\caption{Quantitative evaluation by cumulative error curves for scan-to-mesh distances among local refinement stages.}
	  \label{fig:supple_expr_geo_curves_stages}
\end{figure}

\begin{figure}[ht]
	\centering
	\includegraphics[width=0.9\linewidth]{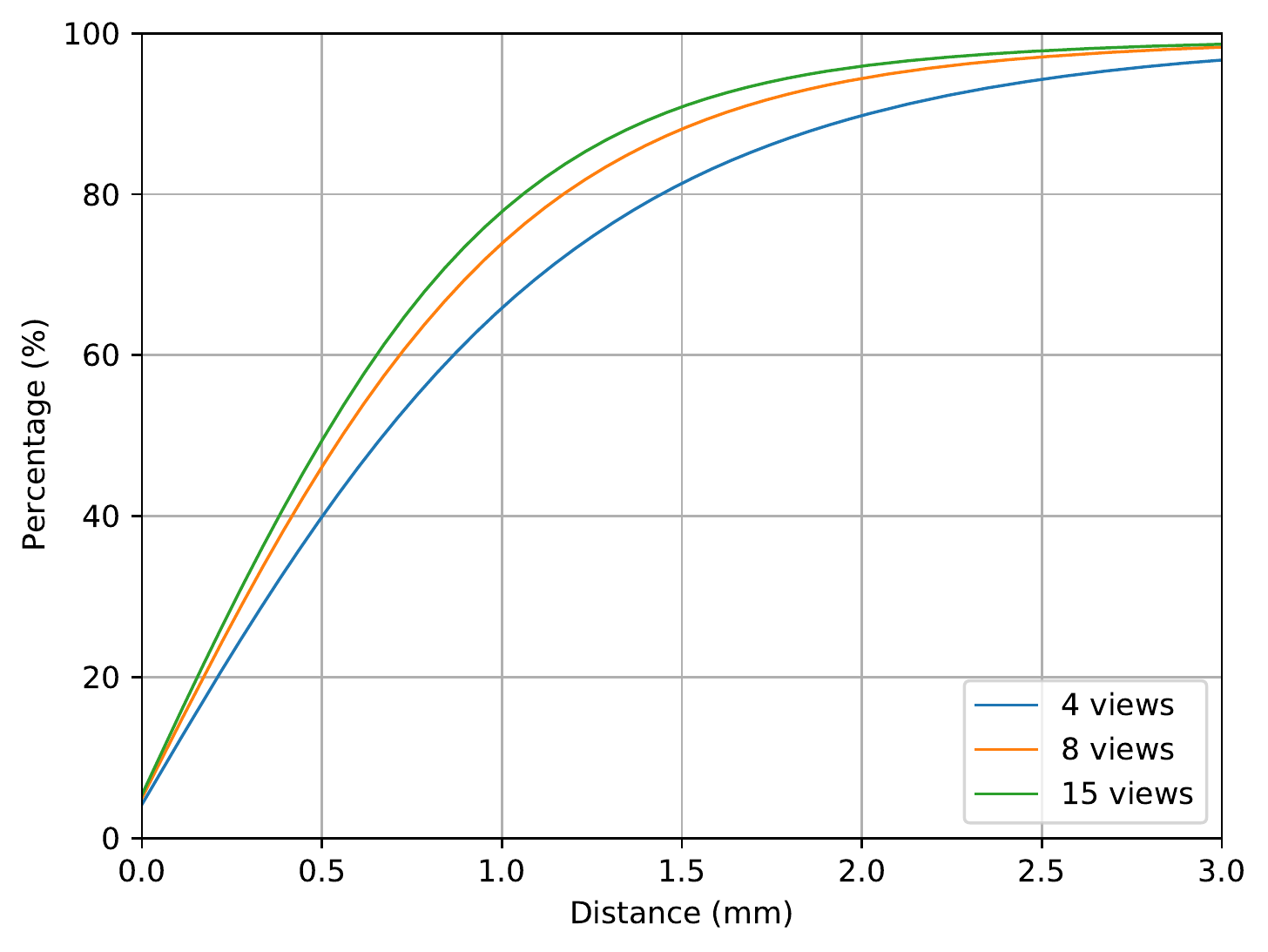}
	\caption{Quantitative evaluation by cumulative error curves for scan-to-mesh distances among various numbers of views.}
	  \label{fig:supple_expr_geo_curves_views}
\end{figure}

\begin{figure}[ht]
	\centering
	\includegraphics[width=0.4\linewidth]{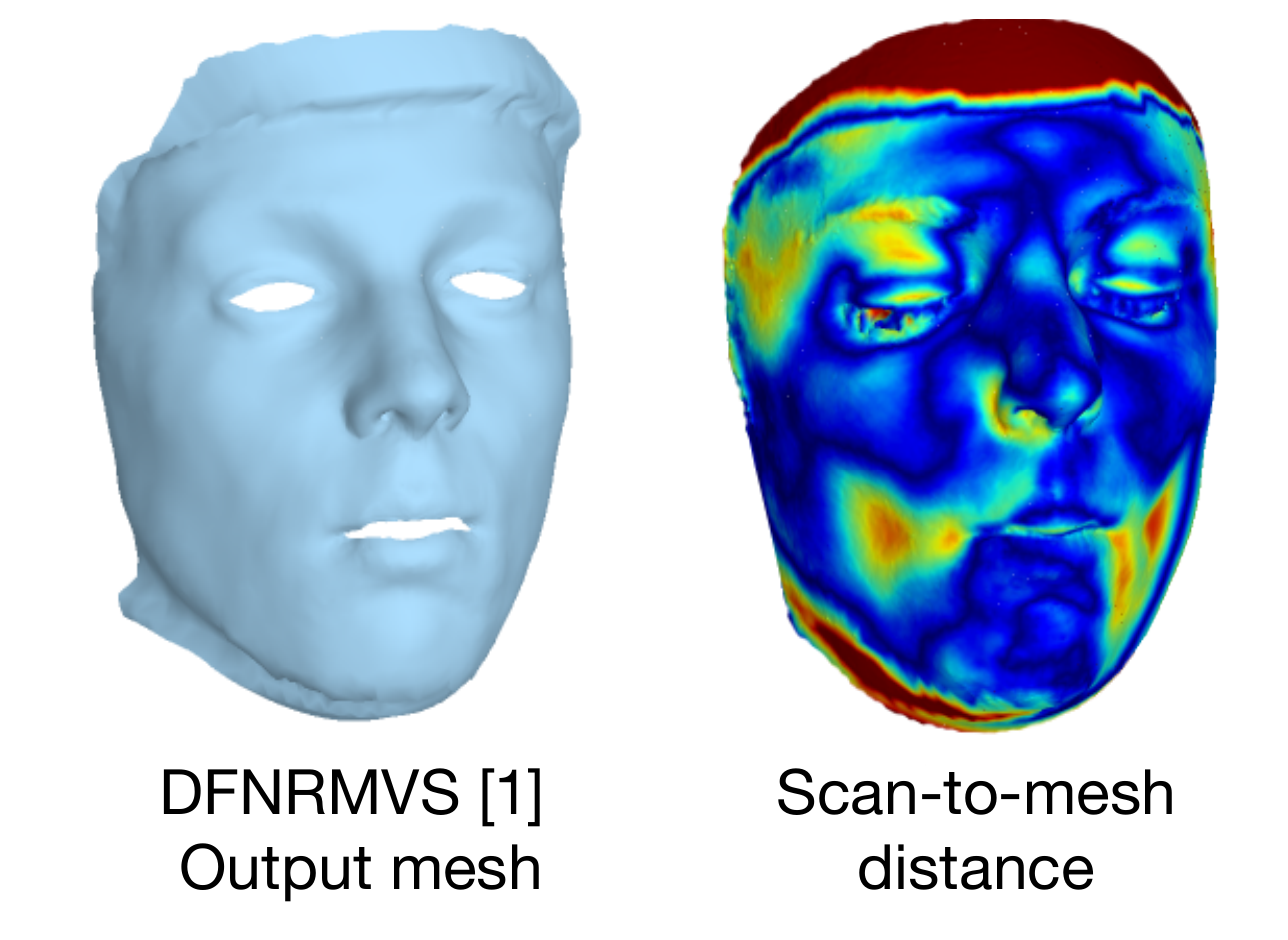}
	\caption{Example results from DFNRMVS \cite{bai2020dfnrmvs}.}
	\label{fig:supple_dfnrmvs_example}
\end{figure}

\begin{figure}[]
	\centering
	\includegraphics[width=0.9\linewidth]{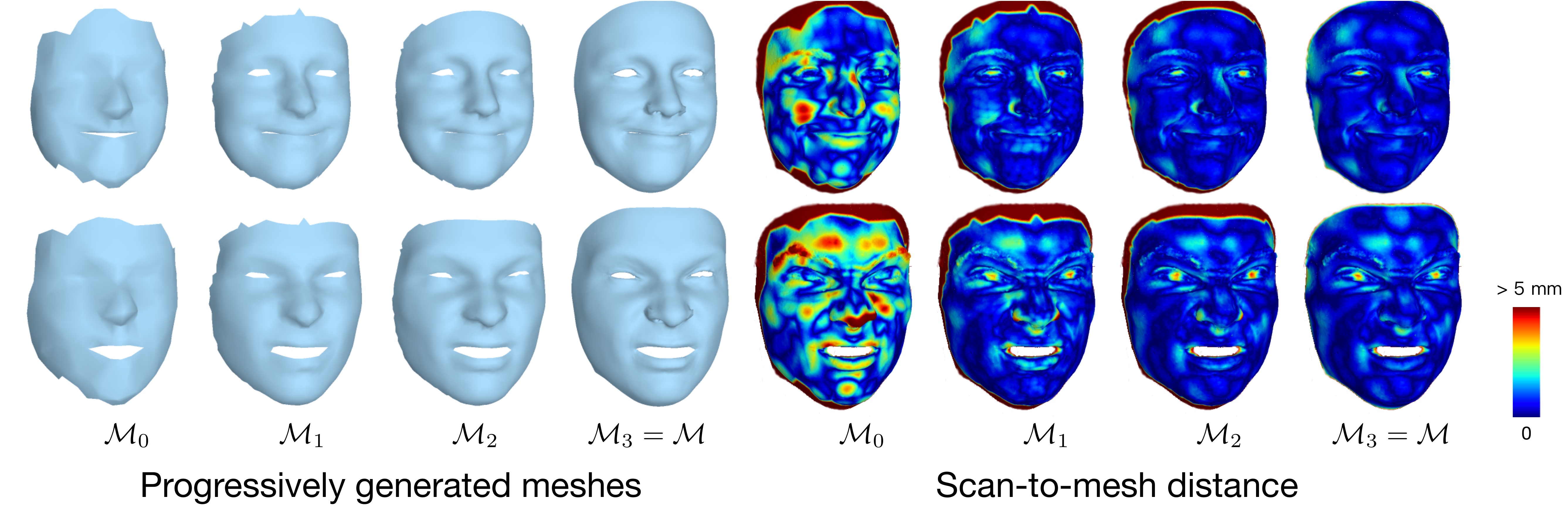}
	\caption{Inferred meshes for global stage $\mathcal{M}_0$ and after upsampling and refinement for each local stage $\mathcal{M}_i$ ($1\leq i \leq 3$).}
	\label{fig:supple_results_per_stage}
\end{figure}

\begin{figure}[ht]
	\centering
	\includegraphics[width=0.9\linewidth]{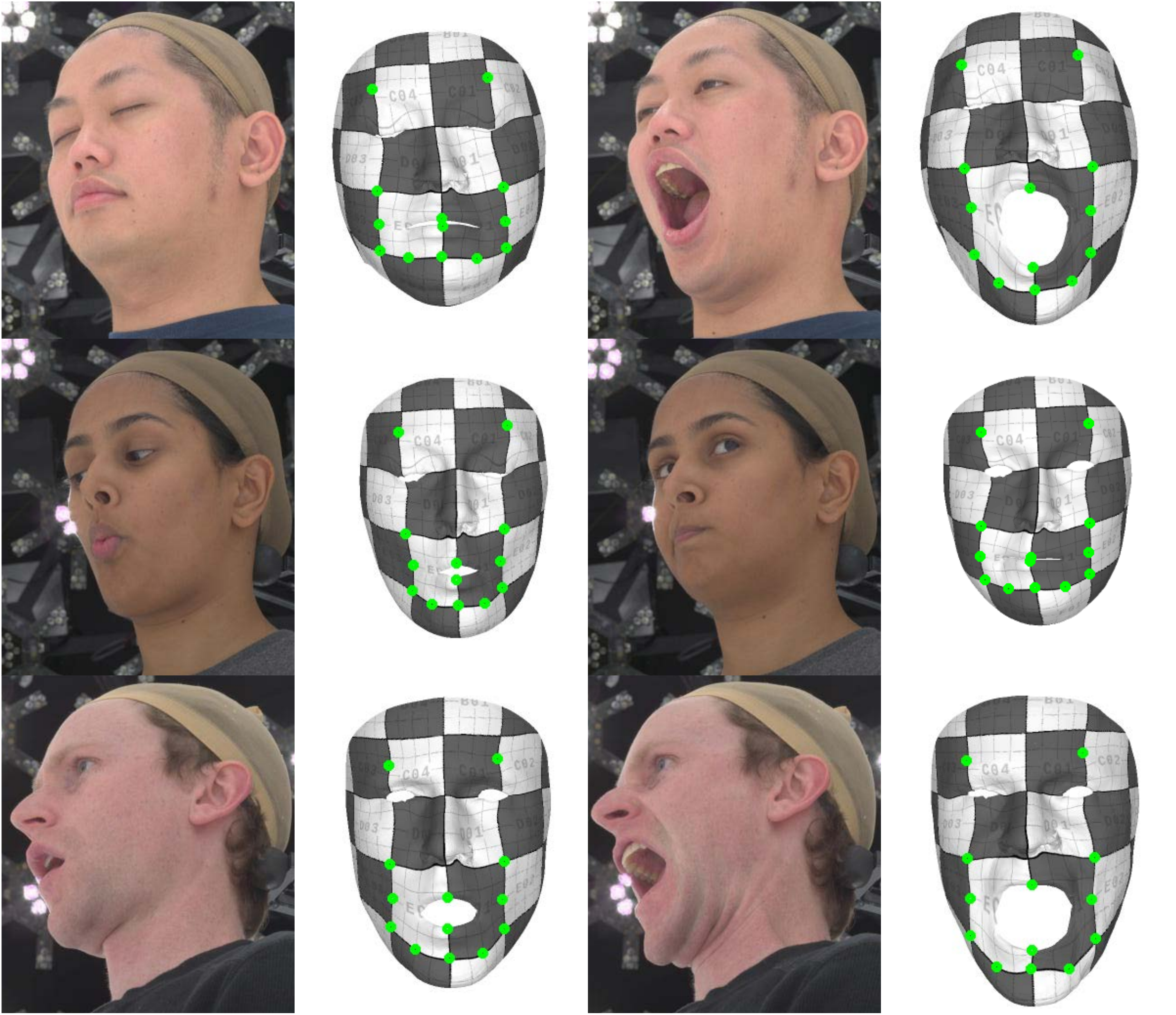}
	\caption{Visualization of cross-subject dense correspondence of the base meshes inferred by \name in a shared checkerboard texture.}
	\label{fig:supple_cross_subject_corresp}
\end{figure}

Our method outperforms (w/o post-processing) the existing methods (w/ and w/o post-processing) in terms of geometric reconstruction quality and the quality of the correspondence.
Note that while the distance of DFNRMVS \cite{bai2020dfnrmvs} is higher than for the 3DMM regression, DFNRMVS \cite{bai2020dfnrmvs} is visually better in most regions.
Their reconstructed meshes tend to have large errors in the forehead and in the jaw areas, as shown in Fig.~\ref{fig:supple_dfnrmvs_example}, due to a different mask definition for their on-the-fly deep photo-metric refinement.
Fig.~5 in the paper shows that our methods produces significantly better reconstructions than DFNRMVS \cite{bai2020dfnrmvs} across the entire face.

\qheading{Additional Qualitative Results.}
We evaluate our trained model on a multi-view video sequence with 8 calibrated and synchronized views, captured at 30 fps.
We apply our progressive mesh generation network in a frame-by-frame manner, without applying any temporal smoothing. 
Fig.~\ref{fig:supple_dynamic_sequence} shows that our base mesh well captures the extreme expressions, and it aligns well with the input images.
Despite being trained on static images only, the resulting reconstruction is temporally stable, as shown in the supplemental video. 
Fig.~\ref{fig:supple_more_results} shows additional base mesh reconstructions for different static multi-view images of varying subjects in different expressions. 
Our method reconstructs the face shape and expression well, closely to the ground truth scans. 
We show more visualizations in the \textit{supplemental video}.

\qheading{Impact of Local Refinements.}
Fig.~\ref{fig:supple_expr_geo_curves_stages} shows the cumulative error curves for scan-to-mesh distances among the local stages.
Given the coarse mesh $\mathcal{M}_0$ as output of the global stage, each local stage successively increases the mesh resolution and refines the vertex locations.
Fig.~\ref{fig:supple_results_per_stage} demonstrates the effect of each local refinement step.
As shown in Fig.~\ref{fig:supple_results_per_stage}, the quality of the reconstructed mesh improves after each local stage, while the scan-to-mesh distance to the scan reduces.
Note that details such as nose corners and lips gradually improve through the local stages.

\qheading{More Ablation on Number of Views.}
Fig.~\ref{fig:supple_expr_geo_curves_views} shows the cumulative error curves for scan-to-mesh distances for networks with different number of input views.

\begin{figure}[ht]
    \centering
    \includegraphics[width=0.9\linewidth]{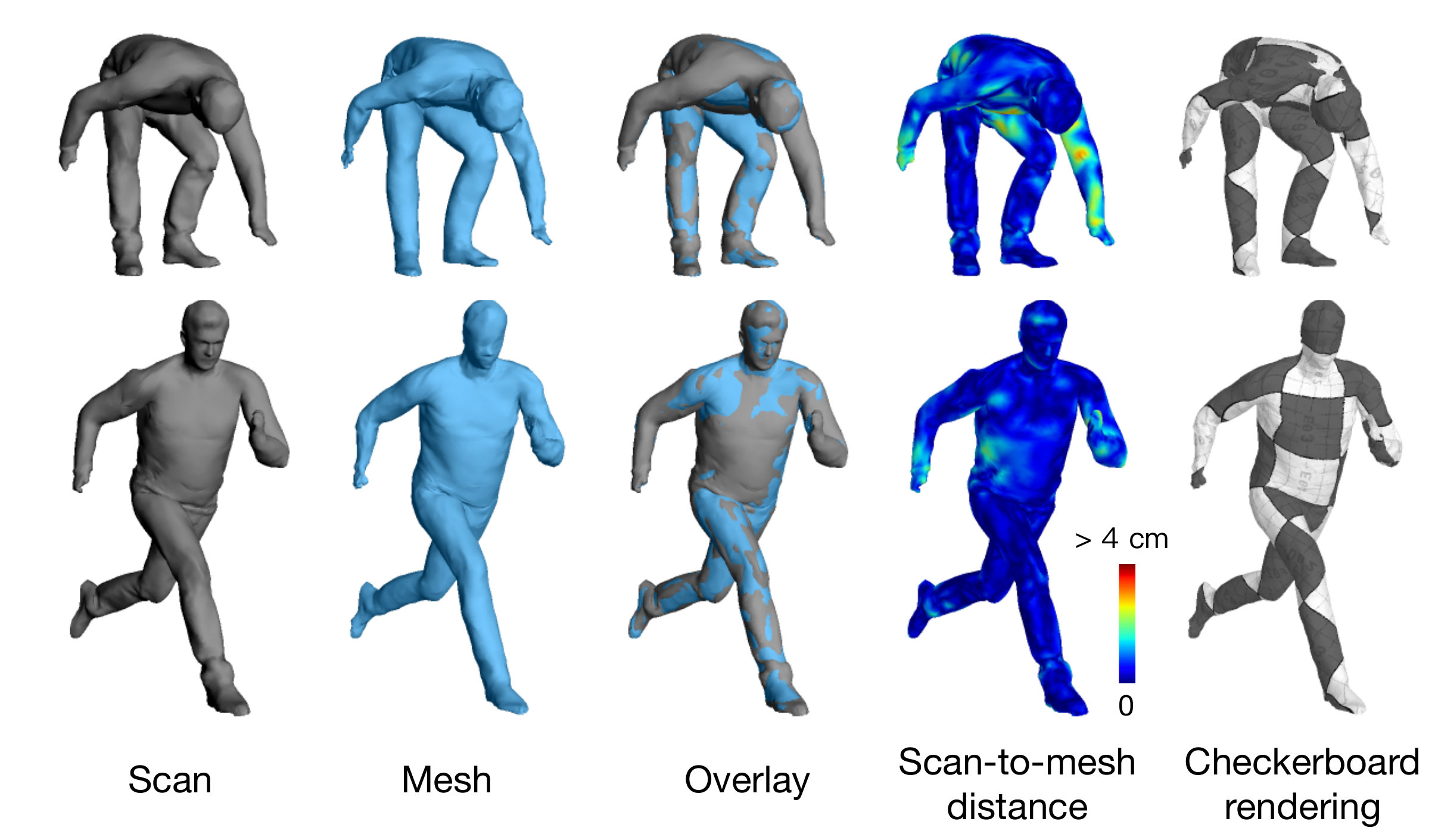}
    \caption{
    Our system can also infer clothed human body surfaces in consistent topology. 
    }
    \label{fig:supple_body_results}
\end{figure}

\qheading{More Results on Appearance and Detail Capture.}
Fig.~\ref{fig:supple_face_details} shows additional results of the appearance enhancement network, which predicts normal displacements and additional albedo and specular maps on top of the predicted base mesh $\mathcal{M}$ (see Fig. 2 of the paper).
Our reconstruction pipeline (i.e. base mesh reconstruction and appearance and detail capture) enables us to reconstruct a 3D face with high-quality assets, 2 to 3 orders of magnitude faster than existing methods, which can readily be used for photorealistic rendering.

\qheading{Results on Clothed Human Body Datasets.}
While we focus on face mesh in correspondence, we find that our method can also predict clothed full body meshes in correspondence. 
We test our method on a dataset of human bodies as shown in Fig.~\ref{fig:supple_body_results}. 
Human bodies are challenging due to large pose variations and occlusions. 
Given the challenging inputs, our methods still outputs detailed geometry which closely fit the ground truth surfaces with small scan-to-mesh distances, shown in Fig.~\ref{fig:supple_body_results}.
Checkerboard projection also shows the accuracy of semantic correspondence among extreme poses. 
The results demonstrate the flexibility of our method for highly articulated and diverse surfaces.

\qheading{Albedo.}
While the input images in our datasets are diffuse albedo images, obtained with polarized lighting and cameras \cite{Ghosh_2011_SIGGRAPH, ma2007rapid}, the results, shown in the paper, indicate that our system can be adapted to non-lightstage setups, e.g. capture system of CoMA \cite{CoMA2018}.
The appearance capture network learns the mapping between albedo images and the details of specular reflectance and fine geometry, as ``image-to-image translation''.
This synthesis is reasonable since the input images contain pore-level details and the outputs are pixel-aligned.
However, imperfect albedo images can potentially contain more information on specularity, which in principle can guide the systhesis network to better recover details.
This is an interesting perspective and we will explore it as future work.

\qheading{The $\mathbb{E}$ Operator.}
Let $B$ be batch size and $N$ be vertex number.
Given a feature volume $\mathbf{L}_{g}$ from the global volumetric feature sampling, the global geometry network (3D ConvNet) predicts a probability volume $\mathbf{C}_{g}$ of size $(B, N, 32, 32, 32)$, whose  $N$-channel is ordered in a predefined vertex order.
Finally the soft arg-max operator $\mathbb{E}$ computes the expectations on $\mathbf{C}_{g}$ \textit{per channel}, and outputs vertices of shape $(B, N, 3)$ corresponding to the predefined order.

\qheading{On Dense Correspondence.}
Dense correspondence across identities and expressions is a challenging task \cite{FLAME2017, bogo2017dynamic}. 
Cross-identity dense correspondence is fundamentally difficult to define beyond significant landmarks, especially in texture-less regions.
The state-of-the-art methods rely on landmarks and propagate the dense correspondence by statistical (3DMM) or physical constraints (Laplacian regularization) in a carefully designed optimization process with manual adjustments.
Cross-expression correspondence, however definable, can be enforced by photometric consistency (optical flow or differentiable rendering). 
Our ground truth datasets utilized all these state-of-the-art strategies and therefore can be regarded as one of the best curated datasets.
With the ``best'' ground truth one can get as now, we trained our network in a supervised manner to the ground truth meshes (same topology) with equal vertex weights.
Measuring the distances to the ground truth (v2v and landmark errors) gives informative and reliable \textit{cross-expression} evaluations on dense correspondence quality.
Furthermore, photometric error visualizations on a shared UV map (as in the main paper) and the stable rendering of reconstructed sequence as in Fig.~\ref{fig:supple_dynamic_sequence} both qualitatively shows high quality of cross-expression correspondence.

However, quantitative evaluating \textit{cross-identity} dense correspondence is by nature difficult.
These two metrics above indirectly measure for cross-subject correspondence. 
Here we show additional visualizations by rendering inferred meshes in a shared checkerboard texture and highlighting some facial landmarks in Fig.~\ref{fig:supple_cross_subject_corresp}.
The meshes inferred by \name preserve dense semantic correspondences across subjects and expressions, as shown by the landmarks and the uniquely textured regions.

\qheading{Implementation Details.}
The appearance enhancement synthesis network uses as similar architecture and losses as proposed by Wang et al.~\cite{wang2018pix2pixHD}.
We train the global generator and 2 multi-scale discriminators at resolution of 512 $\times$ 512.
The main difference is that we extract features from two inputs separately before concatenating them and feeding into the convolutional back-end so that we can better encode useful features correspondingly.
The network is trained using an Adam optimizer with learning rate of $2e-4$ (decayed from 100 epoch) and batch size of 32 on a NVIDIA GeForce GTX 1080 GPU.
For further enhancement, we trained a separate super-resolution network, upsampling attribute maps from 512 to 4K resolution.
We modify the network design from ESRGAN \cite{wang2018esrgan} by expanding the number of Residual in Residual Dense Blocks (RRDB) from 23 to 32, enabling the upsampling capacity from 4$\times$ to 8$\times$ in a single pass.
The super-resolution network is trained with learning rate of $1e-4$ (halved at 50K, 100K, 200K iterations) and batch size of 16 on two NVIDIA GeForce GTX 1080 GPUs.

\begin{figure*}[ht]
	\centering
	\includegraphics[width=0.9\linewidth]{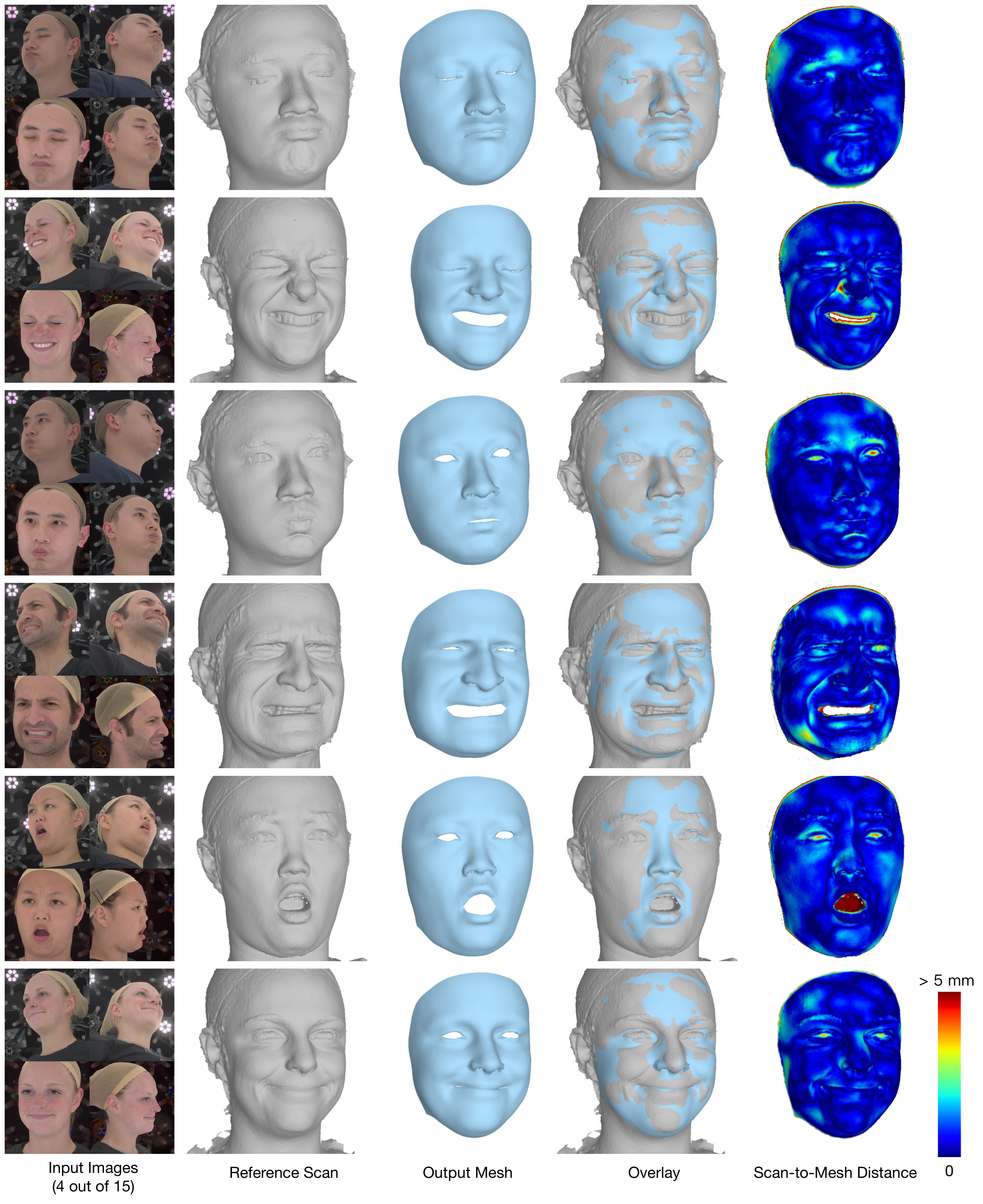}
	\caption{
	More results of reconstructed meshes in dense correspondence. 
	The scan-to-mesh distance is visualized color coded on the reference scan, where red denotes an error above 5 millimeters.}
	\label{fig:supple_more_results}
\end{figure*}

\begin{figure*}[ht]
	\centering
	\includegraphics[width=0.9\linewidth]{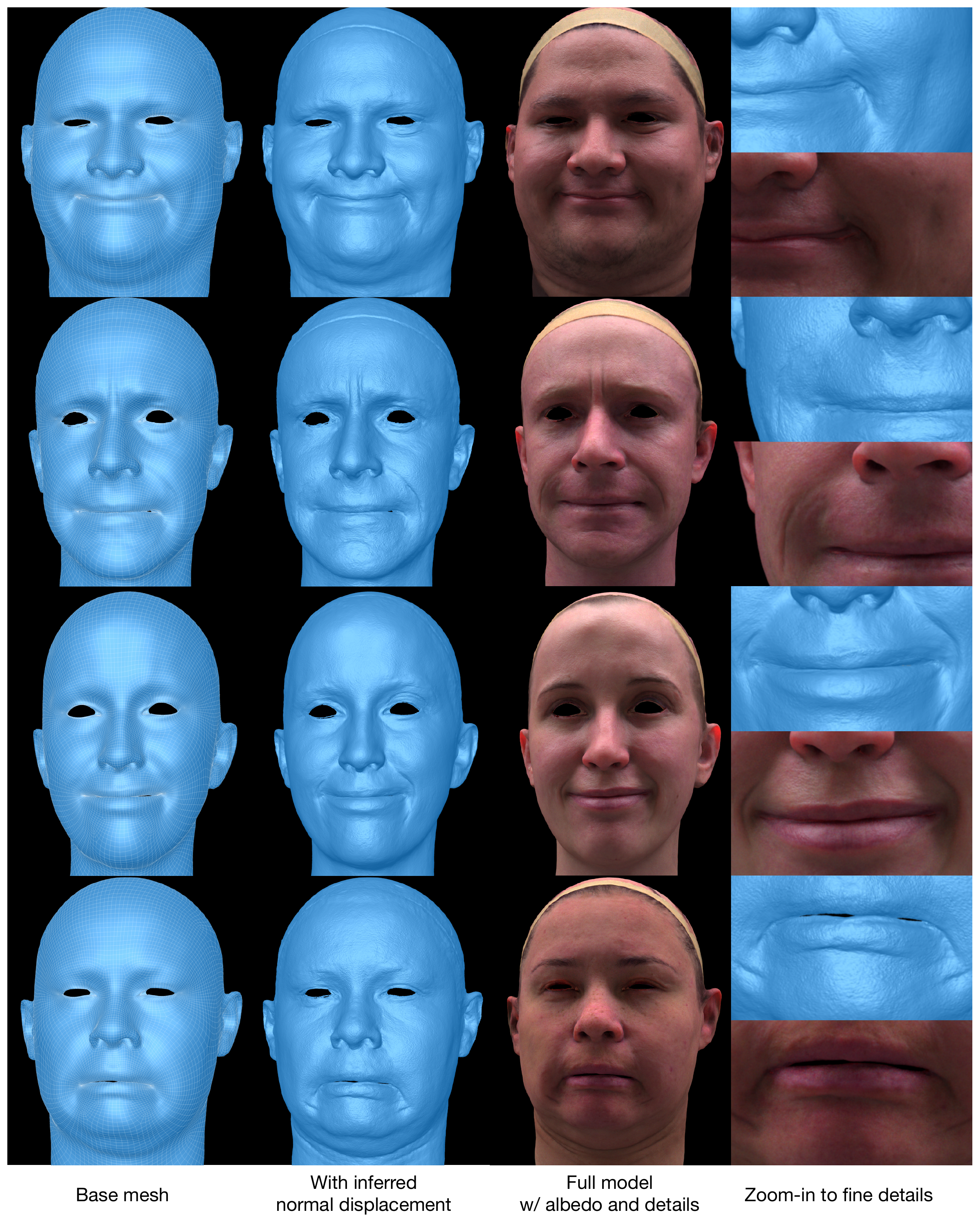}
	\caption{Our method can generate reliable base alignment meshes, on top of which a comprehensive face modeling pipeline can be built. Here we show more rendering with inferred normal displacements and additional albedo and specular maps.}
	\label{fig:supple_face_details}
\end{figure*}

\end{document}